\documentclass[journal]{IEEEtran}
\ifCLASSOPTIONcompsoc
  \usepackage[nocompress]{cite}
\else
  \usepackage{cite}
\fi
\usepackage{ragged2e}
\ifCLASSINFOpdf
  \usepackage[pdftex]{graphicx}
\else
  \usepackage[dvips]{graphicx}
  \DeclareGraphicsExtensions{.eps}
\fi
\usepackage{amsmath}
\usepackage{graphicx,subfigure}
\usepackage{multirow}
\usepackage{booktabs}
\usepackage{algorithm}
\usepackage{algorithmic}
\usepackage{graphicx}
\usepackage{epstopdf}
\usepackage{array}
\usepackage{amsfonts,amssymb}
\ifCLASSOPTIONcompsoc
\usepackage[caption=false,font=footnotesize,labelfont=sf,textfont=sf]{subfig}
\else
\usepackage[caption=false,font=footnotesize]{subfig}
\fi
\usepackage{fixltx2e}
\usepackage{dblfloatfix}
\usepackage{url}
\usepackage{bbm}
\usepackage{amsmath,amsfonts,amssymb,amsthm}
\usepackage{chngcntr}
\newtheorem{ass}{Assumption}
\newtheorem{lemma}{Lemma}

\newtheorem{theorem}{Theorem}
\newtheorem{corollary}{Corollary}
\makeatletter
\newcommand*\bigcdot{\mathpalette\bigcdot@{.5}}
\newcommand*\bigcdot@[2]{\mathbin{\vcenter{\hbox{\scalebox{#2}{$\m@th#1\bullet$}}}}}
\makeatother
\usepackage{color,colortbl}

\begin{document}

\title{MP-FedCL: Multiprototype Federated Contrastive Learning for Edge Intelligence}

\author{Yu Qiao,
        Md. Shirajum Munir,~\IEEEmembership{Member, IEEE},
        Apurba Adhikary,
        Huy Q. Le,
        Avi Deb Raha, \\
        Chaoning Zhang,~\IEEEmembership{Member, IEEE},
        and Choong Seon Hong,~\IEEEmembership{Senior, IEEE}
\IEEEcompsocitemizethanks{

\IEEEcompsocthanksitem Yu Qiao and Chaoning Zhang are with the Department of Artificial Intelligence, School of Computing, Kyung Hee University, Yongin-si 17104, Republic of Korea (email: qiaoyu@khu.ac.kr; chaoningzhang1990@gmail.com).
\IEEEcompsocthanksitem Md. Shirajum Munir is with the School of Cyber Security, Old Dominion University, Suffolk, VA 23435, USA, and also with the Department of Computer Science and Engineering, Kyung Hee University, Yongin-si 17104, Republic of Korea (e-mail: munir@khu.ac.kr).
\IEEEcompsocthanksitem Apurba Adhikary, Huy Q. Le, Avi Deb Raha, and Choong Seon Hong are with the Department of Computer Science and Engineering, School of Computing,
Kyung Hee University, Yongin-si 17104, Republic of Korea (e-mail: apurba@khu.ac.kr; quanghuy69@khu.ac.kr; avi@khu.ac.kr; cshong@khu.ac.kr).
}}

\IEEEtitleabstractindextext{%
\begin{abstract}
\justifying 
Federated learning-assisted edge intelligence enables privacy protection in modern intelligent services. However, not independent and identically distributed (non-IID) distribution among edge clients can impair the local model performance. The existing single prototype-based strategy represents a class by using the mean of the feature space. However, feature spaces are usually not clustered, and a single prototype may not represent a class well. Motivated by this, this paper proposes a multi-prototype federated contrastive learning approach (MP-FedCL) which demonstrates the effectiveness of using a multi-prototype strategy over a single-prototype under non-IID settings, including both label and feature skewness. Specifically, a multi-prototype computation strategy based on \textit{k-means} is first proposed to capture different embedding representations for each class space, using multiple prototypes ($k$ centroids) to represent a class in the embedding space. In each global round, the computed multiple prototypes and their respective model parameters are sent to the edge server for aggregation into a global prototype pool, which is then sent back to all clients to guide their local training. Finally, local training for each client minimizes their own supervised learning tasks and learns from shared prototypes in the global prototype pool through supervised contrastive learning, which encourages them to learn knowledge related to their own class from others and reduces the absorption of unrelated knowledge in each global iteration. Experimental results on MNIST, Digit-5, Office-10, and DomainNet show that our method outperforms multiple baselines, with an average test accuracy improvement of about 4.6\% and 10.4\% under feature and label non-IID distributions, respectively.
\end{abstract}

\begin{IEEEkeywords}
Federated learning, edge intelligence, contrastive learning,  multi-prototype, global prototype pool, label and feature non-IID,  communication efficiency.
\end{IEEEkeywords}}
\maketitle
\IEEEdisplaynontitleabstractindextext
\IEEEpeerreviewmaketitle

\section{Introduction}\label{sec:introduction}
\IEEEPARstart{I}{ncreasingly}, intelligence devices in distributed networks are showing explosive growth, which generates a huge amount of raw data that needs to be processed~\cite{asif2018toward}. Because of the challenges of limitation in network bandwidth or the requirements for transmission delay, the traditional cloud computing paradigm that uploads such big data to a cloud centre for data processing can no longer meet these demands \cite{shi2016edge}. Thanks to the improvements in storage and computing capabilities of edge intelligence devices, most computing tasks can now be completed directly at the edge, making the mobile edge computing (MEC) paradigm the next-generation computing network \cite{deng2020edge}. Further, collecting data from distributed devices poses risks and challenges due to the sensitive nature of a large amount of data, as well as regulations such as the general data protection regulation (GDPR) \cite{magdziarczyk2019right} in Europe. Therefore, as edge devices' storage and computing power continue to grow, coupled with concerns about privacy issues, it becomes more attractive to implement edge intelligence in MEC systems in a distributed manner \cite{wu2022node}. To this end, federated learning (FL), as one application of edge computing in distributed machine learning, is first proposed by \cite{mcmahan2017communication} to simultaneously achieve edge intelligence and address privacy concerns. It trains a global model through the cooperation between local clients and an edge server while keeping the clients' raw data within their respective local environments. In general, the typical federated training process consists of the following four steps \cite{mcmahan2017communication}: (1) the server chooses a certain network architecture such as convolutional neural network (CNN) as the global model to be optimized and sends it to local clients; (2) the clients update the received the model parameters of the global model based on their local data; (3) all clients send their updated model parameters back to the edge server for aggregation; (4) the server averages all the sent parameters as the new global model parameters for the next global round, repeating these four steps until convergence. In this fashion, FL's ability to protect data confidentiality and enable multiple parties to cooperatively train a model makes it a highly promising technology for the future of network intelligence\cite{park2019wireless}.

Nonetheless, a main challenge in FL is that data distribution among clients is usually not independent and identically distributed (non-IID), which can result in reduced effectiveness of FL \cite{kairouz2021advances,li2020federated}. To tackle the issue, existing research works under non-IID scenarios can be mainly divided into two categories in terms of optimization objectives, i.e., typical FL \cite{mcmahan2017communication,zhang2022federated,long2023multi,qiao2023cdfed} and personalized FL \cite{li2020federated, t2020personalized,fallah2020personalized,arivazhagan2019federated,qiao2023framework}. The former objective is to develop a single shared global model that is accurate and efficient, while also capable of adapting to the unique characteristics of each client's data. FedAvg \cite{mcmahan2017communication} is the first FL optimization algorithm to enable efficient training of machine learning models on decentralized data. It achieves this through collaborative training of a shared global model via model parameter transmission among clients in each global round. FedLC \cite{zhang2022federated} adopts a fine-grained calibration strategy for clients' cross-entropy loss to mitigate the bias caused by label distribution skewness among clients in the global model. However, training the global model directly with heterogeneous data from local clients can result in poor generalization abilities to unseen data \cite{zhu2021federated}. In contrast, the latter approach focuses on optimizing local models individually for each client rather than using a shared global model. This is typically achieved by adding a regularization term to the local objective of each client to guide their local training, which enables the models to generalize well to new data. FedProx \cite{li2020federated} proposes to add a local regularization term in the local objective of each client to correct the bias between local models and the global model. PFedMe \cite{t2020personalized} proposes to add an additional term to allow clients to update their local models in different directions without deviating from a global reference point. FedPer \cite{arivazhagan2019federated} proposes a strategy of adding a personalized layer to the base layer and suggests updating only the base layer during the federated training process. Afterwards, clients can update their personalized layer based on their own local data. Additionally, \cite{li2022federated} explores a benchmark for non-IID settings, they divide non-IID settings into five cases, such as label distribution skew, feature distribution skew, quantity skew, etc. Further, as \cite{li2022federated} mentioned, some existing studies \cite{li2020federated,zhang2022federated,long2023multi,tan2022fedproto} cover only one non-IID case, which do not give sufficient evaluations to this challenge. Therefore, to avoid the influence of biased global models and to evaluate non-IID cases as comprehensively as possible, we focus on personalized FL by optimizing the local objective of each local client under the label and feature distribution skewness.

\begin{figure*}[t]
\centering
\subfigure[Local training of client 1.]{
\includegraphics[scale=0.6]{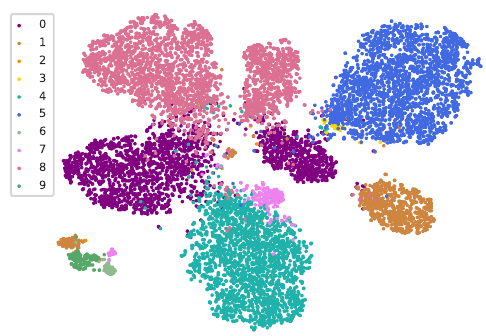}
}
\quad\quad\quad\quad\quad\quad\quad
\subfigure[Local training of client 2.]{\includegraphics[scale=0.6]{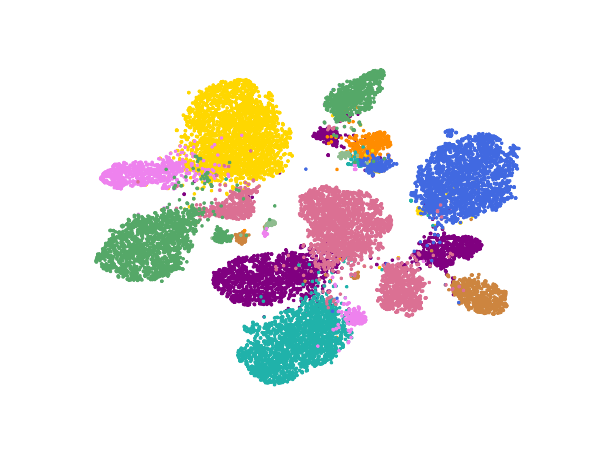}
} \\
\quad
\subfigure[Federated training of client 1.]{\includegraphics[scale=0.6]{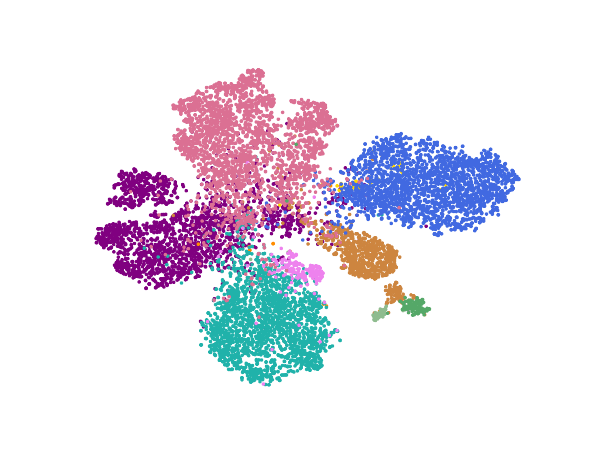}}
\quad\quad\quad\quad\quad\quad\quad
\subfigure[Federated training of client 2.]{\includegraphics[scale=0.6]{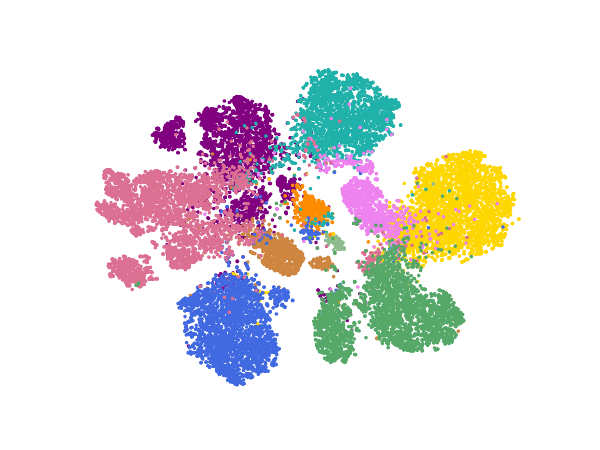}}
\caption{A toy example: T-SNE \cite{van2008visualizing} visualization of the embedding vectors of local clients. The sample distribution for each client follows the Dirichlet distribution \cite{yurochkin2019bayesian} with Dir(0.1). The upper two represent local training, which means no communication with other clients. The below two represent federated training based on FedAvg.}
\label{motivation}
\end{figure*}

Inspired by prototypical networks \cite{snell2017prototypical}, which adopts a single prototype to represent each class by calculating the mean of the class's embedding space. This prototype can serve as an important information carrier to boost the performance of various learning domains, and has been successfully applied in meta-learning \cite{hou2022imbalanced}, multi-task learning \cite{kang2011learning}, and transfer learning \cite{quattoni2008transfer}. There have been some existing works \cite{tan2022fedproto, mu2023fedproc, tan2022federated,qiao2023prototype} introducing the concept of prototypes into FL. FedProto \cite{tan2022fedproto} proposes to reduce communication overhead by only exchanging prototypes between clients and the server, instead of exchanging gradients or model parameters. FedPCL \cite{tan2022federated} proposes to use multiple pre-trained models to extract the features separately, and then they use a projection network to fuse these extracted features in a personalized way while keeping the shared representation compact for efficient communication. These works adopt a single prototype to represent each class and argue that directly averaging the representations from heterogeneous data across clients can effectively capture the embedding representations of each class. 

However, in their approach, they represent each class with a single prototype obtained by averaging over the same class space for each client, which can be considered intuitively incomplete and ambiguous~\cite{huang2021local}. For instance, consider the class of dogs, which includes various breeds differing in size, color, or appearance. Averaging all the dogs within the same class may not adequately capture the diversity and distinctive features present in different dog breeds. This limitation could lead to a less expressive and less discriminative representation for each class, potentially impacting the overall model's performance in scenarios with intra-class variations. Here, we consider a toy example as shown in Figure \ref{motivation} to illustrate this intuition. The figure showcases the embeddings in the class space during local training (the upper two subfigures) and federated training (the bottom two subfigures) under the setting of data heterogeneity among clients. It becomes evident from the visualization that there is considerable separation and diversity across the embedding class space in both training scenarios. In local training, each client refines the model using its own data, leading to distinct representations of each class. Similarly, in federated training, where multiple clients collaboratively train the global model, the embedding space still demonstrates significant variations and differentiation. This observation highlights the non-trivial nature of using a single prototype to sufficiently capture the entire embedding space, whether during local or federated training.

Motivated by the above intuition, we introduce MP-FedCL, a strategy that improves the classification performance of FL in scenarios where there is skewness in the distribution of labels and features. The proposed approach uses a contrastive learning scheme, which employs multiple prototypes to represent each class, thereby learning a more differentiated prototype representation in the embedding space for each class. {\color{black} Here, we are inspired by some recent studies~\cite{rippel2015metric,deuschel2021multi,huang2021local}, which use \textit{k-means} to cluster features in their methods. For instance,  \cite{rippel2015metric} employed \textit{k-means} clustering to learn a density representation within the embedding space. Moreover, in recent studies by~\cite{deuschel2021multi,huang2021local}, a method for calculating multiple prototypes based on \textit{k-means} clustering was proposed and achieved good results in image classification tasks. These findings highlight the effectiveness of using \textit{k-means} clustering to perform feature-based clustering. However, it has not been validated in the FL setting, and our scheme is the first new attempt to introduce the concept of multi-prototype into FL.} Specifically, our proposed strategy first applies the \textit {k-means} clustering algorithm to multiple prototypes calculation, in which each client can calculate their own multiple prototypes for each class. {\color{black} Note that considering the ever-increasing computing capabilities of local clients~\cite{wu2022node,liu2022adaptive,beitollahi2023federated} and privacy issues caused by transferring raw data features~\cite{zhang2022embedded,alazab2021federated,anaissi2021intelligent}, we apply the \textit{k-means} clustering on the feature space on the local side.} Due to the natural clustering properties of \textit {k-means}, the output (i.e. $k$ centroids) of \textit {k-means} clustering algorithm can be viewed as the calculated multiple prototypes for that class. These calculated prototypes from various distributed clients are then sent to the edge server for aggregation as a global prototype pool. The global prototype pool is a combination of multiple prototypes from each client, and it can be updated during training in each global round. To regularize individual local training, we reformulate the local objective of each client in a contrastive learning manner by conducting any supervised learning task (e.g. a cross-entropy loss) and a contrastive learning task. The goal of the contrastive learning task is that prototypes in the global prototype pool and local representations belonging to the same class are pulled together while simultaneously pushing apart those prototypes from different classes. Note that the \textit{k-means} clustering algorithm is conducted on the local side, and the prototype is a one-dimensional vector of low-dimensional samples that are naturally small and privacy-preserving, which does not incur excessive communication costs or raise privacy concerns compared to the model parameters. To the best of our knowledge, we are the first to present multi-prototype learning in FL. The preliminary version of this work has been published in \cite{qiao2023framework} where we design a multi-prototype-based federated training framework for model inference in the last global iteration based on the typical federated training process. The major differences between the current work and \cite{qiao2023framework} are the addition of the global prototype pool based on a contrastive learning strategy, the modification of the global iteration process, and the exploration of the feature distribution skewness. Our main contributions to this paper are as follows:
\begin{itemize}
\item[$\bullet$] We introduce a \textit{k-means}-based multi-prototype federated contrastive learning (MP-FedCL) framework, designed to capture both intra-class and inter-device information. The former can be achieved through the multi-prototype strategy, while the latter can be accomplished by the global prototype pool for information exchange.
\item[$\bullet$] We reformulate the loss function for each client to perform both supervised learning and contrastive learning tasks. This strategy can encourage each client to learn their own supervised learning task, while also learning from the global prototype pool.
\item[$\bullet$] We demonstrate that our proposed strategy outperforms several baselines on multiple benchmark datasets regarding test accuracy and communication efficiency, with improvements of about 4.6\% and 10.4\% under feature and label skewness, respectively.
\end{itemize}

The remainder of this article is organized as follows. Related work on federated learning and prototype learning is presented in Section \ref{relatedwork}. The system model and problem formulation are provided in Section \ref{sec:SMPF}. The strategy for multi-prototype computation and aggregation, as well as the model inference, are presented in Section \ref{sec:MPFL}. Experimental results are provided in Section \ref{sec:experiments}. Finally, conclusions are drawn in Section \ref{sec:conclusion}.

\section{Related Work}
\label{relatedwork}
In this section, we first review the existing works to deal with FL challenges in Section \ref{rel_FL}, including the statistical and system heterogeneity, and communication efficiency. Then, we briefly review some works that apply prototype learning to FL in Section \ref{rel_PL}, followed by a schematic diagram of our proposed multi-prototype FL, which will be explained in detail in the next section.

\vspace{-0.2cm}
\subsection{Federated Learning}
\label{rel_FL}
One of the key challenges in FL is the distribution of training data across multiple clients, which is usually statistically heterogeneous (also known as the non-IID issue). This heterogeneity can limit the effectiveness and performance of FL. Many existing works \cite{wang2020tackling, karimireddy2020scaffold, luping2019cmfl, yao2019federated, Nader2020Adaptive,caldas2018expanding} are dedicated to improving communication efficiency under the challenge of statistical heterogeneity. Other works \cite{chai2021fedat, xie2019asynchronous, chen2020asynchronous, park2021handling, wang2022asynchronous} are mainly from the perspective of system heterogeneity, dealing with communication efficiency issues under the system heterogeneity.

To tackle the system heterogeneity, FedAT \cite{chai2021fedat} introduces an asynchronous layer in which clients are grouped according to their system-specific capabilities to avoid the straggler problem, thus reducing the total number of communication rounds. Unlike typical federated training processes, which are usually implemented with synchronous approaches and can cause stragglers and heterogeneous latency, FedAsync \cite{xie2019asynchronous} combines asynchronous training with federated training to tackle that issue. To solve the lag or dropout problems of distributed edge devices during federated training, ASO-Fed \cite{chen2020asynchronous} presents an online federated learning strategy, in which edge devices use continuous streaming local data for online learning and an edge server aggregated model parameters from clients in an asynchronous manner. Sageflow \cite{park2021handling} proposes a robust FL framework to cope with both stragglers and adversaries problems. In this framework, clients are grouped and weighted according to their staleness (i.e., arrival delay). Then, entropy-based filtering and loss-weighted averaging are applied within each group to defend against attacks from malicious adversaries. Another asynchronous FL framework in a wireless network environment is proposed in \cite{wang2022asynchronous}. The framework aims to adapt to environments with heterogeneous edge devices, communication environments, and learning tasks by considering possible delays in local training and uploading local model parameters, as well as the freshness between received models. However, most of these works do not consider the statistical heterogeneity which is the major challenge in FL.

\begin{figure}[t]
\centering
\subfigure[Single prototype.]
{\includegraphics[height=4cm,width=4.2cm]{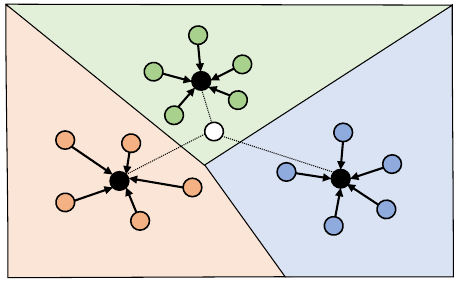}}
\subfigure[Multiple prototypes.]
{\includegraphics[height=4cm,width=4.45cm]{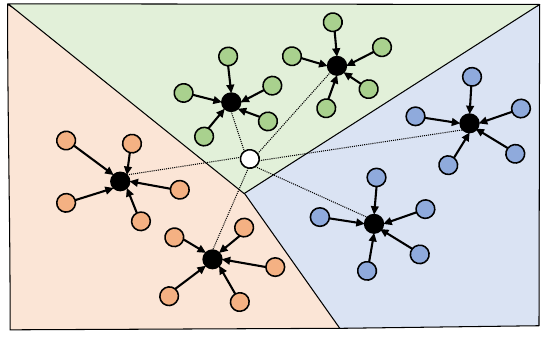}}
\caption{A diagram illustrates two strategies for model inference: the single prototype-based (left) and multi-prototype-based (right) approaches. In the diagram, black solid circles represent prototypes, while white solid circles represent query classes.}
\label{prototype_based_Inference}
\vspace{-0.1cm}
\end{figure}

To address the statistical heterogeneity, FedNova \cite{wang2020tackling} suggests that different clients can perform a different number of local steps when updating their shared global model with their local private data. SCAFFOLD\cite{karimireddy2020scaffold} introduces two control variables which contain the updated direction information of the global model and local models to overcome the gradient difference and effectively alleviate client drift problems. CMFL \cite{luping2019cmfl} designs a feedback mechanism that can reflect the updated trend of the global model. Each client in the system checks whether it is consistent with the update trend of the global model before uploading its model updates to the edge server, otherwise, it does not upload. This strategy of uploading only information related to model improvement to the server greatly reduces communication overhead. FedMMD \cite{yao2019federated} employs a two-stream model to extract a more generalized representation by minimizing the maximum mean difference (MMD) loss, which is a measure of the distance between two data distributions. This approach can accelerate the convergence rate and reduce communication rounds. AFD \cite{Nader2020Adaptive} proposes a dynamical sub-model parameters selection method, in which clients can update their models using a sub-model rather than the whole global model parameters. This sub-model selection strategy is performed by maintaining an independent activation score map for each client. At each round, the server sends a different sub-model for selected clients, and then clients update their respective score maps according to their own local loss function. A similar approach is also considered in Fed-Dropout \cite{caldas2018expanding}, which adopts a lossy compression way for server-to-client communication while allowing clients to update their models using sub-models of the global model, further reducing communication overhead. However, most of these works do not consider the scenario where clients in FL are under heterogeneous feature distribution.

\begin{figure}[t]
\centering
{\includegraphics[scale=0.70]{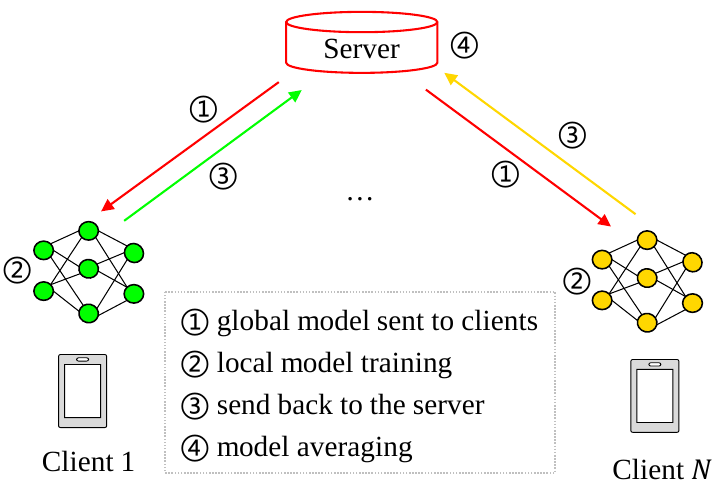}}
\caption{An overview of federated training framework. In the beginning of each global round, the server first sends the global model to clients (step.1). Each local client then performs local training based on its own training data (step. 2). Next, after the local training is finished, each client send the updated model parameters to the server for aggregation (step. 3). Finally, the model parameters are averaged at the server (step. 4).}
\label{overview_framwork}
\vspace{-0.4cm}
\end{figure}

\vspace{-0.2cm}
\subsection{Prototype Learning}
\label{rel_PL}
Prototype learning is first proposed by prototype networks \cite{snell2017prototypical} in few-shot learning. Its design idea is to use a single prototype to represent one class, where the single prototype is calculated by averaging the embedding vectors within the same class space. Prototype learning has received significant progress in various tasks such as image classification \cite{michieli2021continual}, video processing \cite{xue2022dynamic}, and natural language processing \cite{wieting2015towards} areas. In image classification tasks \cite{michieli2021continual}, a class is represented as a single prototype by computing the mean of the feature vectors of that class. In video processing \cite{xue2022dynamic}, prototypes are obtained by calculating the average feature over different timestamps. In natural language processing tasks \cite{wieting2015towards}, taking the average of word embeddings can yield a prototype representation for a sentence. Further, both few-shot learning and FL are based on the scenario of training with a small amount of data: the clients do not have enough data to train their own models. Recently, there have been various successful works using prototypes for federated optimization in computer vision tasks. In FedProto \cite{tan2022fedproto}, the authors propose to reduce communication overhead by only exchanging prototypes between clients and the server instead of exchanging gradients or model parameters. However, their work does not validate in a more general heterogeneous environment setting such as Dirichlet distribution \cite{yurochkin2019bayesian}. An optimized prototype-based FL is proposed in \cite{michieli2021prototype} by using margins of prototypical representations learned from distributed heterogeneous data to calculate the deviations of clients and applying these deviations through an attention mechanism to boost model performance. FedProc \cite{mu2023fedproc} introduces that global prototypes in the server can be used as a guideline to correct clients' training in local updates, and they use a contrastive loss to pull each class to be close to the corresponding global prototypes while pushing away from other global prototypes. FedPCL~\cite{tan2022federated} proposes a strategy based on contrastive learning that uses single prototype exchange instead of gradient communication for efficient communication. MOON~\cite{li2021model} is designed based on model-level contrastive learning by comparing the representations between the global model and local models. However, most of them focus on each client's individual learning process while disregarding the collaborative contributions from other clients. Moreover, most of the above-mentioned works adopt to represent the same class using a single prototype, which may fail to capture discriminative embedding representations by naively taking the mean of the feature space \cite{li2020adaptive,li2021adaptive}.

\begin{table}[!t]
\caption{Summary of Notations.}
\label{tab: notations}
\centering
\begin{tabular}{|c||l|}
\hline
Notation & Description\\
\hline
$\mathcal{D}_i$ & Heterogeneous dataset of client $i$\\
$\boldsymbol {x}_i$ & Feature space of client $i$\\
$y_i$ & Corresponding label in the feature space of client $i$\\
$D_i$ & Size of dataset $\mathcal{D}_i$\\
$\omega$ & Shared model parameters of global model \\
$\mathcal F(\cdot)$ & Shared global model\\
$[C]$ & Label space set\\
$C$ & Number of label space\\
$p_j(\cdot)$ & Probability of sample being classified as the $j$-th class\\
$\mathbbm{1}(\cdot)$ & Indicator function\\
${f_i} (\cdot)$ & Empirical risk of client $i$ with one-hot encoded labels\\
$\mathcal{L}_i(\cdot)$ & Local loss of client $i$\\
$[N]$ & Set of clients\\
$N$ & Number of clients\\
$\mathcal{L}(\cdot)$ & Global loss across all clients\\
$\eta$ & Learning rate\\
$\nabla \mathcal{L}_i(\bigcdot)$ & Loss gradient of client $i$\\
$\mathcal L_S$ & Supervised learning loss\\
$\mathcal L_R$ & Regularization term\\
$c$ & A local representation of one client\\
$\mathbb{U}_j$ & Aggregated global prototype set belonging to $j$-th class\\
$u_i$ & One instance in global prototype set $\mathbb{U}_j$\\
$\mathbb{U}$ & Aggregated global prototype pool from all clients\\
$f_{e}(\cdot)$ & Feature extraction layers\\
$f_{c}(\cdot)$ & Decision-making layers\\
$v_{i,j}$ & Embedding space of client $i$ belong to class $j$\\
{\color{black} $u_{i,j}$} & Output of clustering($v_{i,j}$)\\
{\color{black} $|u_{i,j}|$} & {\color{black} Number of $u_{i,j}$ for class $j$ of $i$-th client} \\ 
$K$ & Number of clusters\\
$\overline{U}_j$ & Averaged value of global prototype set $\mathbb{U}_j$\\
$N_i$ & Number of instance belonging to $\mathbb{U}_j$\\
$\bigcdot$ & Inner (dot) product\\
$\tau$ & Temperature hyperparameter \\
$P(y)$ & Set of labels distinct from $y$ \\
$|P(y)|$ & Size of $P(y)$ \\
$p$ & $p \in P(y)$ \\
$A_p$ & Size of labels distinct from $p$ \\
$\hat y$ & Predicted label \\
$\Vert \bigcdot \Vert$ & $\ell_2$-norm of a vector \\
$E$ & Number of local epoch\\
$B$ & Batch size\\
$T$ & Number of global communication rounds\\
$\mathbb E(\cdot)$ & Expectation\\
$\zeta$ & $\zeta$-Lipschitz \\
$L_1$ & $L_1$-smooth \\
$\delta$ & $\delta$-local dissimilar \\
$G$ & Stochastic gradient of each client $i$ is bounded by $G$ \\
$L_2$ & $L_2$-Lipschitz continuous \\
\hline
\end{tabular}
\vspace{-0.1cm}
\end{table}

Therefore, different from the single prototype learning paradigm used in these works, we propose to use multiple prototypes to represent each class and adopt multi-prototype-based contrastive learning to capture intra-class differences and inter-class similarities. Here, we briefly describe the core part of the model inference process in the single-prototype-based approach and our proposed multi-prototype-based approach, as shown in Figure \ref{prototype_based_Inference}. Taking the multi-prototype strategy, as illustrated in the right subfigure of Figure \ref{prototype_based_Inference}, as an example, the model inference stage involves two main processes: distance calculation and decision-making. When a query class is introduced to the network during inference, these two processes are executed to determine the most appropriate prototype for the query class. Firstly, the network calculates distances between the new query class and the multiple prototypes associated with each existing class in the class space. These prototypes represent the diverse representations learned from different clients during the FL process. Secondly, the classification decision for the new query class is made based on the shortest distance to any of the prototypes of that specific class. The network assigns the new query class to the class whose prototype exhibits the closest similarity in the embedding space. Note that the multi-prototype concept used in the model inference strategy is also used in our contrastive learning-based model training process. However, the training process differs from the multi-prototype-based model inference process. During the contrastive learning-based model training, the objective is to optimize the query to be as close as possible to multiple prototypes belonging to the same class space while simultaneously ensuring that it remains far away from all other prototypes belonging to class spaces different from its own. The contrastive learning technique aims to enhance the discriminative power of the model by encouraging similar representations for data points belonging to the same class and pushing apart those from different classes. This way, the model learns to create well-separated and informative embeddings for each class, which in turn is expected to benefit the subsequent inference process. The details of the multi-prototype calculation and multi-prototype-based model training and inference will be illustrated in the following sections. Table \ref{tab: notations} presents a summary of the notations used in this manuscript.

\begin{figure*}[t]
\centering
\includegraphics[width=1.0\textwidth]
{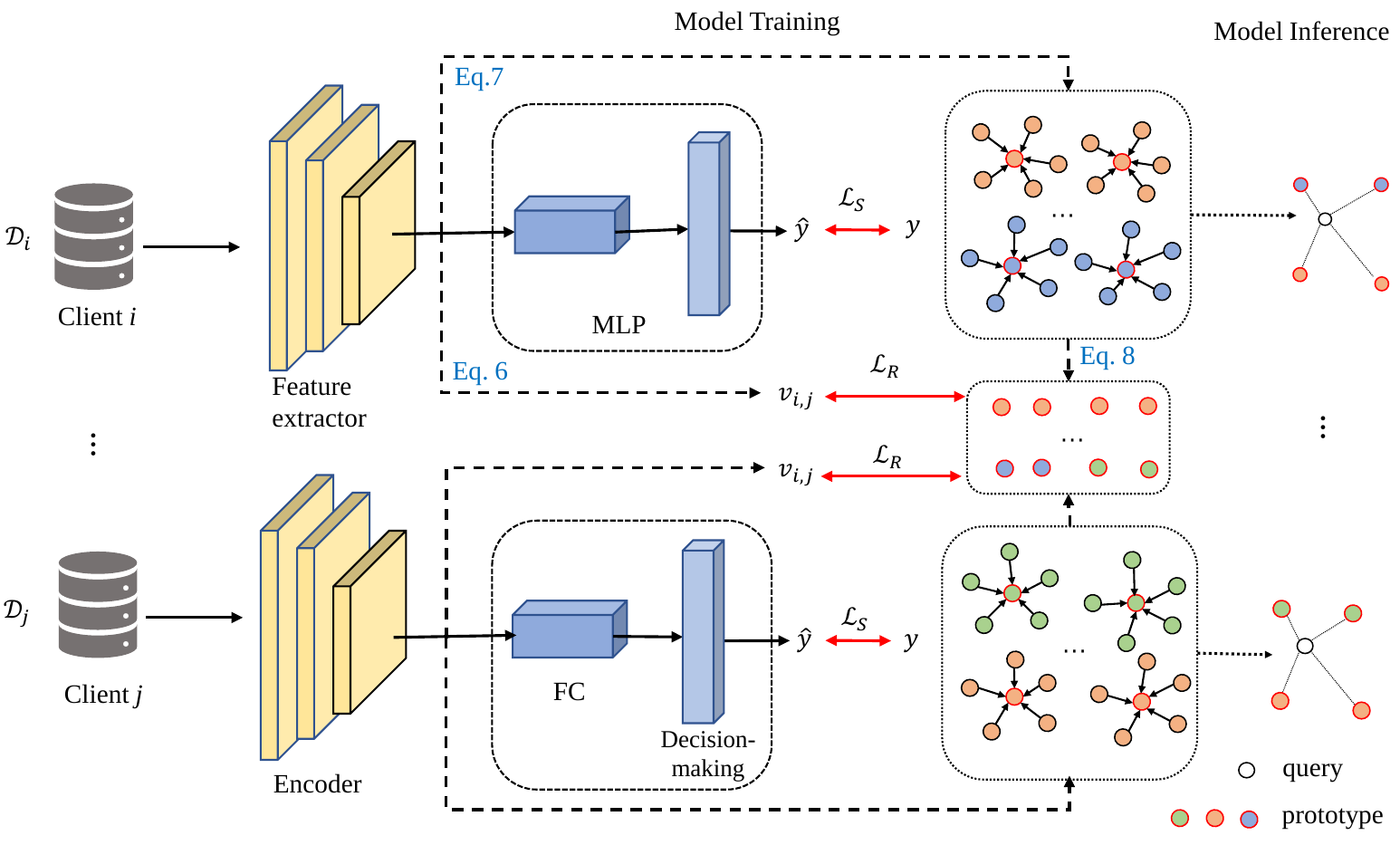}
\caption{An overview of the proposed multi-prototype based FL framework (here illustrated with $K$ = 2 prototypes). In each global round, clients not only transmit their model parameters but also the calculated multiple prototypes by \textit{k-means} to the server for aggregation. The aggregated prototypes set (the aggregated prototypes set is also called the global prototype pool) is then sent back to each client along with updated model parameters for the next global iteration. Finally, the clients update their local models by minimizing the loss of typical supervised learning loss $\mathcal L_s$ and the distance between the global prototype pool and local representations $\mathcal L_R$ in a contrastive way, repeating the above process until convergence.}
\label{System_model}
\end{figure*}

\section{System Model and Problem Formulation}
\label{sec:SMPF}
In this section, we first introduce the key elements behind FL, including the system model in Section \ref{FLM}  and the local training optimization algorithm for local training in Section \ref{SO}. Then, our optimization problem is formulated in Section \ref{PF}.
\vspace{-0.2cm}
\subsection{Federated Learning Model}
\label{FLM}
The essence of the federated learning strategy is to train a model through the collaboration of distributed clients based on their local data, which serves the purpose of protecting data privacy. The overview of the FL framework is shown in Figure \ref{overview_framwork}. The training process can be summarized as follows:

The typical process of federated learning is based on the setting where each client $i$ has a heterogeneous and privacy-sensitive dataset, denoted by $\mathcal{D}_i= {(\boldsymbol {x}_i, y_i)}$, of size $D_i$, where $\boldsymbol {x}_i$ and $y_i$ represent the feature space and corresponding label of the $i$-th client, respectively. The goal is to coordinate the collaboration between clients and the edge server to train a shared model $\mathcal F(\omega; \boldsymbol {x}_i)$ for each client. The empirical risk (e.g., cross-entropy loss) of the client $i$ with one-hot encoded labels can be defined as follows \cite{mu2023fedproc}:
\begin{equation} \label{one_hot}
   {f_i} (\omega) = -\sum_{j=1}^{C}\mathbbm{1}_{y = j}\log p_j(\mathcal F(\omega; \boldsymbol {x}_i); y_i),
\end{equation}
where $\mathbbm{1}(\cdot)$ is the indicator function, $\omega$ is the shared model parameters of global model, $C$ is the number of classes belonging to label space $[C] = \{1,..,C\}$, and $p_j(\mathcal F(\omega; \boldsymbol {x}_i); y_i)$ denotes the probability of data sample $(\boldsymbol {x}_i, y_i)$ being classified as the $j$-th class. In addition, the local training of each client is to minimize the local loss $\mathcal{L}_i$ as follows:
\begin{equation} \label{local_loss}
   {\mathcal{L}_i} (\omega) = \frac {1}{D_i} \sum_{i \in \mathcal{D}_i} f_i(\omega).
\end{equation}

The global objective is to minimize the loss function across heterogeneous clients as follows:
\begin{equation} \label{global_loss}
    \mathop {\arg\min}_{\omega} {\mathcal{L}} (\omega) = \sum_{i \in [N]} \frac {D_i}{\sum_{i \in [N]}D_i} {\mathcal{L}_i} (\omega),
\end{equation}
where $[N]$ denotes the set of distributed clients with $[N] = \{1,...,N\}$.

\subsection{SGD Optimization}
\label{SO}
As the first FL optimization algorithm, FedAvg requires multiple global iterations during the training process. Most subsequent works \cite{zhang2022federated,long2023multi,qiao2023cdfed,t2020personalized, fallah2020personalized, arivazhagan2019federated} follow this training framework, including our work. In each global iteration process, each selected client participates in training and performs local stochastic gradient descent (SGD) to optimize its local objective:
\begin{equation} \label{sgd}
 \omega_{t+1}  =  \omega_{t} - \eta \nabla \mathcal{L}_i(\omega_{t}),
\end{equation}
where $\eta$ is the learning rate, $\nabla \mathcal{L}_i(\bigcdot)$ is the loss gradient of client $i$, and $\omega_{t}$ is the updated result of the global model in the previous round.

\vspace{-0.2cm}

\subsection{Problem Formulation}
\label{PF}
The system model is shown in Figure \ref{System_model}, and the illustration of the transmission of model parameters between clients and the edge server is omitted for simplicity. In each global iteration, the server needs to not only receive model parameters from clients and perform model parameters averaging, but also receive prototype knowledge from clients and aggregate them (the strategy for prototype aggregation will be explained in Section \ref{GPA}). Finally, the aggregated model parameters and global prototype pool are returned to local clients participating in the training, while the next iteration begins until convergence. In summary, during the model training phase, clients $i$ and $j$ with heterogeneous datasets $\mathcal{D}_i$ and $\mathcal{D}_j$, respectively, need to simultaneously transmit their own model parameters and their own prototypes based on the features extracted from the feature extraction layers (a.k.a the encoder) to be used in the next global round. Note that the prototypes are the output after clustering, rather than directly averaging the output of the feature extractor layers. The MLP includes the fully connected (FC) layers and a decision-making layer (a.k.a the classifier). In order to extract better features, the former can be an ordinary convolutional layer or a pre-trained network, and the latter is used to map the output of the former from one latent space to another for further representation learning. Each client aims to minimize the typical supervised learning loss $\mathcal L_S$, while also minimizing the distance between their local representations and the global prototype pool for samples belonging to the same class space, and maximizing the distance for samples not belonging to the same class space. This can be denoted as $\mathcal L_R$. During the model inference stage, the clients can use their individual updated local representations to compare with the updated global prototype pool for model inference. We have marked these entities in the system model and given the basic prototype calculation process.

Specifically, motivated by prototype learning in FL and the observation in their works \cite{li2020adaptive, li2021adaptive}, 
the goal of this paper is first to learn multi-prototype representations for each class space through the federated training process, and then perform the final model inference based on these prototypes. Formally,
\begin{equation} \label{base_model_infer}
     \mathop{\arg\min}_{j} \|c -  u_i\|, \quad u_i \in \mathbb{U}_j, \quad j \in [C],
\end{equation}
where $c$ is a local representation of one client, $\mathbb{U}_j$ is defined to aggregate the multiple prototypes for each client belonging to $j$-th class, and $u_i$ denotes one instance in corresponding aggregated global prototype set $\mathbb{U}_j$. Finally, the prediction can be made by measuring the distance between one local representation $c$ and each aggregated prototypes set $\mathbb{U}_j$ and then choosing the $j$-th label with the smallest distance as the final prediction.

\section{Multi-Prototype Federated Learning}
\label{sec:MPFL}
In this section, we design an MP-FedCL algorithm to improve the performance of federated training. In Section \ref{MPC}, we give the method for calculating multiple prototypes, which is used to compute multiple prototypes in the embedding space for each class based on \textit{k-means} clustering, thus obtaining a relatively full representation for each class. Then, a strategy for multiple prototypes aggregation is proposed in Section \ref{GPA} to collect all class-related knowledge shared by all clients. The objective function of the proposed algorithm is presented in Section \ref{Obj}. During the model inference stage, predictions are made based on the distance from the prototypes rather than through a classifier in Section \ref{Infer}.

\vspace{-0.2cm}
\subsection{Multi-Prototype Calculation}\label{MPC}
The feature extraction layer and the decision-making layer in the MLP are usually two core parts of the deep learning model. The former is mainly used to extract feature information from the input space, and the latter makes the final prediction decision based on the learned feature information. For any client $i$, we denote its feature extraction layers and the MLP as $f_{e}(\omega_{e}; \boldsymbol {x}_i)$ and $f_{c}(\omega_ {c}; \boldsymbol {x}_i)$, then the embedding space of its $j$-th class instance can be calculated as:
\begin{equation} \label{feature_extrac}
 v_{i,j} = \{f_{e}(\omega_{e}; \boldsymbol {x}_i) \,\,|\,\, \boldsymbol {x}_i \in \mathcal D_{i,j} \},
\end{equation}
 where $\mathcal D_{i,j}$ is a set of $\mathcal{D}_i$ that belongs to the $j$-th class, and $v_{i,j}$ is the embedding space of $j$-th class.

\begin{table}[t]
\centering
\caption{Comparison between model parameters and corresponding multi-prototype on various datasets.}
\label{tab:com_paras}
\resizebox{\linewidth}{!}{%
\begin{tabular}{|c||c|c|c|} 
\hline
/ & Model Parameters & Multi-Prototype & \textit{Ratio} \\ 
\hline
MNIST & 798,474 & 5,120 & $\approx$ 0.006 \\ 
\hline
Digit-5 & 133,898 & 10,240 & $\approx$ 0.076 \\ 
\hline
Office-10 & 133,898 & 10,240  & $\approx$ 0.076 \\ 
\hline
DomainNet & 133,898 & 7,680  & $\approx$ 0.057 \\ 
\hline
\end{tabular}%
}
\vspace{-0.2cm}
\end{table}

In order to calculate multiple prototypes for each class, we iteratively cluster $v_{i,j}$ into $k$ clusters based on \textit{k-means} algorithm, which is an effective unsupervised algorithm for clustering and has been proven to converge to at least a local optimum after a small number of iterations \cite{geron2022hands}. 
Based on the standard iteration of \textit{k-means}, similarly we also randomly select the $k$ centroids in the first iteration. We then calculate the centroid to which each sample in $\mathcal D_{i,j}$ should belong to, and repeat this process until the centroid does not change or changes only slightly. The multiple prototypes used in our method are defined as centroids obtained through \textit{k-means} clustering. Specifically, we apply \textit{k-means} clustering to the embeddings of each class, and each resulting centroid is considered a prototype for that class in the embedding space. Thus, multiple prototypes of class $j$ can be defined as follows:
 \begin{equation} \label{cluster}
     u^{k}_{i,j} = \textit {Clustering}(v_{i,j}), \quad k \in \{1, ...,K\},
 \end{equation}
where $K$ is the number of clusters, the output of \textit {Clustering(·)} for the $i$-th client of the $j$-th class is denoted as {\color{black} $u_{i,j}$ where $u_{i,j} = \{u^{k}_{i,j} \mid k\in \{1, ...,K\}\}$}. {\color{black}Note that because our clustering algorithm is executed on the client side, we need to communicate prototypes with the server, but compared to the complete model parameters, the additional communication cost of prototype communication is very small~\cite{he2020group,tan2022fedproto,lou2023decentralized}. The model parameters and the corresponding multi-prototypes, and their ratios are shown in the table \ref{tab:com_paras}. The \textit{Ratio} in the Table \ref{tab:com_paras} is defined as the proportion of multi-prototype to corresponding model parameters, and its result indicates that multi-prototype communication accounts for only a small fraction (less than 0.1) of model parameters. Note that the model architecture and the selection of $K$ for multi-prototype are discussed in detail in Section~\ref{sec:experiments}.}

\subsection{Multi-Prototype Aggregation}
\label{GPA}
 The goal of multi-prototype learning is to reduce the distance of embedding space between local prototypes and the corresponding prototypes from the global prototype pool. After receiving the clustering results as computed in Eq. \ref{cluster}, the server groups these prototypes (also denotes $k$ centroids) by class as a global prototype pool. {\color{black} This pool is denoted as $\mathbb{U}$ = \{$\mathbb{U}_1$,$\mathbb{U}_2$,...,$\mathbb{U}_C$\}, where each group contains prototypes from the same class but from different clients.} Formally,
\begin{equation} \label{aggregation}
     \mathbb{U}_j = \{u_{i,j}^{k}  \,|\, i \in [N] \}, k \in \{1,...,K\},
\end{equation}
where $\mathbb{U}_j$ denotes the set of clients that own multiple prototypes of class $j$. Through this aggregation mechanism, {\color{black} the prototypes are grouped according to their respective class labels. The resulting global pool summarizes all class-related knowledge shared by all clients, allowing us to effectively utilize information from various clients while maintaining class-wise separation.}

After the aggregation process, the server sends back the global prototype pool to local clients, which is used to guide their local training in the next global round.  {\color{black} However, some clients' classes may be missing~\cite{mu2023fedproc,li2021fedrs,zhou2022efficient} or underrepresented (i.e., classes with very few samples)~\cite{servetnyk2020unsupervised,fraboni2021clustered,shuai2022balancefl}. In such cases, the global prototype pool can be utilized to fill in the missing classes and ensure that prototypes for all classes have the same number of prototypes for the learning process. Here, inspired by the single prototype padding approach used in various tasks~\cite{pal2023extreme,tan2022federated,zhou2023pmr}, we introduce a multi-prototype-based padding procedure to achieve a balanced representation of all class prototypes per client. By introducing this prototype padding process, each client is guaranteed to have a consistent and balanced set of prototypes for all classes, regardless of their respective data distribution. Specifically, for the averaging process, the prototypes from different clients are matched based on their respective class labels, which means that prototypes belonging to the same class but originating from different clients are grouped together. Next, we perform the averaging of prototypes by summing up all the prototypes belonging to a particular group and then dividing the sum by the total number of prototypes in that group. Formally,} 
 \begin{equation} \label{padding}
     \overline{U}_j = \frac{1}{N_i}\sum_{i=1}^{N_i}u_i, u_i \in \mathbb{U}_j,
 \end{equation}
where $\overline{U}_j$ represents the averaged value belonging to global prototype set $\mathbb{U}_j$, $N_i$ denotes the number of instance belonging to $\mathbb{U}_j$.

{\color{black}
Subsequently, we implement the prototype padding procedure to ensure that each client has a consistent and balanced set of $K$ prototypes for every class, which can be formulated as follows:
\begin{equation}
\forall k \in \{1, ..., K\}, \quad u^{k}_{i,j} =
\begin{cases}
    \overline{U}_j, & \text{if } |u_{i,j}| < K \\
    u^{k}_{i,j}, & \text{if } |u_{i,j}| = K
\end{cases},
\label{apply_padding}
\end{equation}
where $|u_{i,j}|$ is the number of $u_{i,j}$ for class $j$ of $i$-th client. We traverse all $K$ possible prototypes $u^{k}_{i,j}$ for class $j$ of the $i$-th client. For each $k$, we check the number of prototypes $|u_{i,j}|$ currently available for class $j$ of the $i$-th client. If $|u_{i,j}|$ is less than $K$, we apply prototype padding by replacing $u^{k}_{i,j}$ with the averaged prototype $\overline{U}_j$. This ensures that each client has $K$ prototypes for each class, even in cases where some prototypes might be missing or underrepresented. On the other hand, if $|u_{i,j}|$ equals $K$, it indicates that the client already has the required number of prototypes for class $j$, and no padding is needed. In this case, we keep the $k$-th prototype $u^{k}_{i,j}$ unchanged.}

\vspace{-0.2cm}
\subsection{Objective Function}
\label{Obj}
As shown in Figure \ref{System_model}, our proposed network architecture consists of two loss functions. The first one, $\mathcal L_S$, is the loss of typical supervised learning tasks, which can be computed using Eq. \ref{global_loss}. The second one, $\mathcal L_R$, is our proposed supervised contrastive loss term. After receiving the global prototype pool from the server, the objective of local clients, $\mathcal L_R$, is to align their local representations with the corresponding prototype in the global prototype pool, while simultaneously pushing away dissimilar prototypes from themselves. As such, each client can benefit from other clients. We define the supervised contrastive loss \cite{khosla2020supervised} as: 
\begin{equation} \label{L_R}
 \small
    \mathcal L_R = \sum_{i \in \mathcal{D}_i}\frac{-1}{KN} \sum_{i=1}^{N} \sum_{k=1}^{K}\frac{1}{|P(y)|}\sum_{p\in P(y)}\log\frac{\exp(v_i \bigcdot u_{p}^{k} / \tau)}{\sum_{a \in [C]}\exp(v_i \bigcdot u_{a}^{k} /\tau)},
 \end{equation}
where the $\bigcdot$ symbol denotes the inner (dot) product, $\tau$ is a scalar temperature hyperparameter (the smaller the temperature coefficient, the more focused it is on difficult samples.), $v_i$ represents the local embedding of client $i$, $P(y)$ is the set of labels distinct from $y$ and the size of $P(y)$ is $|P(y)|$, and $k$ denotes one element in a certain global prototype set and the size of one set is $K$. 
 
 Therefore, the global objective for the network can be formulated as:
\begin{equation} \label{obj:global_loss}
     \mathcal L(\omega_i  \,|\, {\mathcal{D}_i}) = \mathcal L_S(\mathcal F(\omega_i; \boldsymbol x_i); y_i) + \mathcal L_R.
\end{equation}
A more detailed model training pseudocode for MP-FedCL is shown in Algorithm \ref{alg: MP-FedCL_training}. The inputs of this algorithm are heterogeneous datasets and some training parameters. After the network is initialized, the federated training process is performed from line 3 to line 20. The multiple prototypes calculation and aggregation are dealt with in lines 19 and 7, respectively. In each global round, the local representations for each client are calculated in line 13. The supervised learning task for them is computed in line 14, and the regularization term is calculated in line 15. After the stochastic gradient descent in the local clients has been performed in line 16, each client then sends their own updated model parameters and calculated multiple prototypes in line 20 back to the server for model parameters aggregation in line 9 and multiple prototypes aggregation in line 7, repeating the above iteration process for $T$ rounds until convergence.

\begin{algorithm}[htpb] 
    \caption{MP-FedCL for federated training} 
    \label{alg: MP-FedCL_training} 
    \begin{algorithmic}[1] 
        \REQUIRE ~~ \\
        Dataset $\mathcal{D}_i(\boldsymbol{x}_i, y_i)$ of each client, learning rate $\eta$, \\
        Number of local clients $N$, number of local epoch $E$,\\
        Number of global communication rounds $T$, \\
        Number of clusters $k$. \\
    \end{algorithmic}
    \quad \textbf{Edge server executes:}
    \begin{algorithmic}[1]
        \STATE {Initialize $\omega^0$}
        \STATE {Initialize global prototype set $\mathbb{U}$} for all classes
        \FOR{ $t$ = 1, 2, ..., $T$ } 
            \FOR{ $i$ = 0, 1,..., $N$ \textbf{in parallel}}
                \STATE Send global model $\omega^t$ 
                and global prototype set $\mathbb{U}$ to client $i$
                \STATE {$\omega_i^t , u^{k}_{i,j} \gets \textbf{LocalUpdate}(\omega_i^t, {\mathbb{U}})$}
                \STATE ${\mathbb{U}} \gets \{\{u_{i,j}^{k}  \,|\, i \in [N], {j \in [C]} \}, k \in \{1,...,K\}\}$
            \ENDFOR
            \STATE {$\omega^{t+1} \gets \sum_{i=1}^{N} \frac{|\mathcal{D}_i|}{|\mathcal{D}|}\omega_i^t$}
        \ENDFOR \\
    \textbf {LocalUpdate($\omega_i^t$, {$\mathbb{U}$}):}
    \FOR{ each local epoch }
    \FOR{ each batch ($\boldsymbol{x}_i$; $y_i$) of $\mathcal{D}_i$ }
        \STATE { $v_{i,j} \gets f_{e^i}(\omega_{e^i}^t; \boldsymbol{x}_i)$}
        \STATE $\mathcal L_S \gets CrossEntropyLoss(\mathcal F(\omega_i; \boldsymbol{x}_i); y_i))$.
        \STATE Calculate $\mathcal L_R$ according to Eq. \ref{L_R}
        \STATE {$\omega_i^t \gets \omega_i^t - \eta\nabla(\mathcal L_S + \mathcal L_R)$}
    \ENDFOR
    \ENDFOR
    \STATE  {$u^{k}_{i,j} \gets \textit {Clustering}(v_{i,j})$} 
    \RETURN $\omega_i^t$, $u^{k}_{i,j}$ \\
    \end{algorithmic}
\end{algorithm}

\begin{algorithm}[t] 
    \caption{MP-FedCL for model inference} 
    \label{alg: MP-FedCL_inference} 
    \begin{algorithmic}[1] 
        \REQUIRE Test dataset of each client \\
    \end{algorithmic}
    \begin{algorithmic}[1]
    \FOR{ each sample $i$ in testing dataset}
        \FOR{ each class $j$ in $\mathbb{U}$}
            \FOR{ each instance $k$ in class $j$}
                \STATE Compute the $\ell_2$ distance between $f_{e}(\omega_{e}; \boldsymbol {x}_i)$ \\
                and each instance in $\mathbb{U}_j$
            \ENDFOR
            \STATE Choose the smallest distance as a candidate-predicted \\
            label for class $j$  according to Eq. \ref{model_infer}
        \ENDFOR
        \STATE Collect all candidate-predicted labels
        \STATE Make final predictions based on the smallest candidate prediction label
    \ENDFOR
    \end{algorithmic}
\end{algorithm}

\vspace{-0.2cm}
\subsection{Model Inference}
\label{Infer}
Based on the findings of the survey in \cite{luo2021no}, it is revealed that the lower test accuracy of models in FL environments can be attributed primarily to the later layers of the model. Specifically, the classifier's predictions have the greatest impact on the model's accuracy. As a result of this finding, we propose an innovative approach that utilizes the output before the decision-making layer in the model for making predictions instead of relying solely on the classifier. This new approach can be expressed mathematically as a reformulation of Eq. \ref{base_model_infer}:
\begin{equation} \label{model_infer}
    \hat y = \mathop {\arg\min}_{j} \|f_{e}(\omega_{e}; \boldsymbol {x})  -  u_i\|,  \quad u_i \in \mathbb{U}_j 
\end{equation}
where $\hat y$ is the predicted label, $f_{e}(\omega_{e}; \boldsymbol {x})$ is the output of feature extraction layers (this symbol is originally denoted as $c$ in Eq. \ref{base_model_infer}, i.e, the output of feature extraction layers is represented as the local representation), and $\Vert \bigcdot \Vert$ denotes the $\ell_2$-norm of a vector. The prediction can be made by measuring the $\ell_2$ distance between the local representation $f_{e}(\omega_{e}; \boldsymbol {x})$ and the aggregated prototypes set $\mathbb{U}_j$ of $j$-th class. A more detailed model inference pseudocode for MP-FedCL is shown in Algorithm \ref{alg: MP-FedCL_inference}. For each client, each sample in its test dataset first calculates the distance to each instance of an aggregated global prototype set in line 4 and then chooses the smallest distance as a candidate-predicted label in line 6. Later, all the candidate-predicted labels are collected in line 8. Finally, the prediction can be made based on the smallest candidate-predicted label in line 9. We also provide a convergence analysis of MP-FedCL in Appendix \ref{appex:CA}.

\subsection{{\color{black}Complexity Analysis}}
{\color{black}
Since Algorithm \ref{alg: MP-FedCL_training} in MP-FedCL involves many similar global iterations, we analyze the time complexity of only one such iteration for simplicity. In Algorithm \ref{alg: MP-FedCL_training}, each global iteration mainly consists of the following steps: communication round, local model training, and \textit{k-means} clustering. The algorithm executes a total of $T$ global communication rounds. In each round, the global model $\omega^t$ and the global prototype set $\mathbb{U}$ are sent to each of the $N$ clients in parallel. Therefore, the time complexity for each round can be considered as $\mathcal{O}(1)$. For the local model training process, each client performs local updates for a fixed number of local epochs $E$. For the convenience of analysis, let us consider a FC neural network where each layer has the same number of parameters, denoted as $M$. Without loss of generality, we assume that the data samples of each client are the same $D$, given that the batch size is expressed as $B$, the total number of iterations required to complete one local iteration is $D/B$. In addition, since the computation in each layer can be viewed as a matrix-vector multiplication (note that the matrix calculation mainly involves weight parameters, we ignore bias parameters calculation for simplicity), the time complexity for forward propagation of the FC network in local model training can be expressed as $\mathcal{O}(E\cdot D / B \cdot L_{layer} \cdot M^2)$, where $L_{layer}$ represents the number of layers in the FC network. Overall, the time complexity for forward propagation can be simplified as $\mathcal{O}(D M^2)$; here, certain variables $E$, $B$, and $L_{layer}$ can be considered as constants since $E$, $B$, and $L_{layer}$ $\ll$ $M$ hold typically. Therefore, the time complexity for local model training is $\mathcal{O}(DM^2 + DM^3)$, where the time complexity for back propagation is $\mathcal{O}(DM^3)$. After local model training, each client performs \textit {k-means} clustering to update its embedding space $v_{i,j}$ for each class. The time complexity for \textit{k-means} algorithm is typically $\mathcal{O}(E \cdot I \cdot K \cdot D \cdot C \cdot M)$, where $E$ denotes the number of local epochs, $I$ is the number of iterations, $K$ is the number of clusters, $D$ is the number of data points, $C$ is the number of classes, and $M$ is the dimensionality of the output before the decision-making layer (here, the output is $M$ because it is assumed that parameters of each layer are $M$). Similarly, certain variables like $E$, $I$, $K$, and $C$ can be considered as constants since $E$, $I$, $K$, $C$ $\ll$ $M$, then the time complexity for the proposed multi-prototype calculation can be simplified to $\mathcal{O}(DM)$. Compared with local model training, the time complexity for multi-prototype calculation is relatively low. For Algorithm \ref{alg: MP-FedCL_inference}, after the embedding space is calculated, for a certain test sample, the time complexity for model inference is $\mathcal{O}(K \cdot C)$. We assume the number of test samples for each client is $D'$, then the total time complexity for model inference is linear as $\mathcal{O}(D')$ since $K$ and $C$ can be considered as constants.}

\section{Experiments}
\vspace{0.1cm}
\label{sec:experiments}
We now present the experimental results of the proposed multi-prototype-based federated learning strategy. We implement MP-FedCL on different datasets, and models and compare with the most commonly used baselines including  Local, FedAvg \cite{mcmahan2017communication}, FedProx \cite{li2020federated}, and FedProto \cite{tan2022fedproto}. We first introduce the datasets and local models in Section \ref{DLM}. Then, the implementation details are provided in Section \ref{ID}. The selection of $K$ for different datasets is discussed in Section \ref{CK}. In Section \ref{AC}, the test accuracy of different baselines under different datasets with different non-IID settings is illustrated. The robustness and communication efficiency comparison are shown in Section \ref{RC} and \ref{CEC}, respectively.

\begin{table}[t]
\centering
\caption{Model Architecture for MNIST.}
\label{tab: model_architecture}
\resizebox{0.8\linewidth}{!}{%
\begin{tabular}{|c||c|c|c|} 
\hline
& Layer & Activation & Value  \\ 
\hline
\multirow{2}{*}{Encoder} 
& FC1& Relu& (28*28, 512)  \\ 
\cline{2-4}
 & FC2 &  Relu  & (512, 512)  \\ 
\hline
\multirow{2}{*}{MLP} & FC1 & Relu  & (512, 256) \\ 
\cline{2-4}
& FC2  & Relu  & (256, 10)  \\
\hline
\end{tabular}%
}
\vspace{-0.2cm}
\end{table}

\vspace{-0.2cm}
\subsection{Datasets and Local Models}
\label{DLM}
We conduct experiments on four popular benchmark datasets: MNIST \cite{lecun1998gradient}, Digit-5\cite{zhou2020learning}, Office-10\cite{gong2012geodesic}, and DomainNet\cite{peng2018synthetic} to verify the potential benefits of multiple-prototype based federated learning for edge network intelligence. \textbf{MNIST} is the handwritten digit recognition dataset. It contains 10 different classes with 60,000 training samples and 10,000 test samples. \textbf{Digit-5} is a collection of images of handwritten digits from the five most popular datasets, including SVHN, USPS, MNIST, MNIST-M and SynthDigits. \textbf{Office-10} consists of images from four different office environments, each containing a distinct set of classes: Amazon (A), Webcam (W), DSLR (D), and Caltech (C). \textbf{DomainNet} is a large-scale, multi-domain image classification dataset. It consists of over 600,000 images from 345 categories, divided into 6 domains: clipart, infograph, painting, quickdraw, real, and sketch. Each domain contains a distinct set of classes and has its own characteristics and challenges.

\begin{table}[!t]
\caption{{Experimental Parameters}.}
\label{tab:parameters}
\centering
\begin{tabular}{|c||c|}
\hline
Parameters & Values\\
\hline
$\text{ {Learning rate} $\eta$}$ & 0.01\\
$\text{ {Learning rate} decay}$ & 0.95\\
$\text{ {Batch size} $B$}$ & 32\\
$\text{ {Local epoch} $E$}$ & 1\\
$\text{ {Temperature} $\tau$}$ & 0.07\\
$\text{ {Dirichlet parameter} $\alpha$}$ & 0.05 (default)\\
$\text{Optimizer}$ & SGD\\
$\text{SGD momentum}$ & 0.5\\
$\text{$K$(MNIST)}$ & 2\\
$\text{$K$(Digit-5)}$ & 4\\
$\text{$K$(Office-10)}$ & 4\\
$\text{$K$(DomainNet)}$ & 3\\
\hline
\end{tabular}
\vspace{-0.2cm}
\end{table}

For local models, a 2-layer encoder network with 2 FC layers and an MLP with 2 FC layers are considered for MNIST, as shown in Table~\ref{tab: model_architecture}. For these datasets that are more complex than MNIST, such as Digit-5, Office-10, and DomainNet, we use ResNet18 \cite{he2016deep}, which has been pre-trained on DomainNet, as the encoder. Please refer to their work \cite{dvornik2020selecting} for more details about the pre-trained model. We employ the same MLP architecture as in MNIST for these datasets. The output dimension of the encoder and the input of the decision-making layer of the MLP network are 512 and 256, respectively. Note that for fair comparisons, all baselines adopt the same network architecture as MP-FedCL, including MLP.

\begin{figure}[t]
\centering
\subfigure{\includegraphics[scale=0.6]{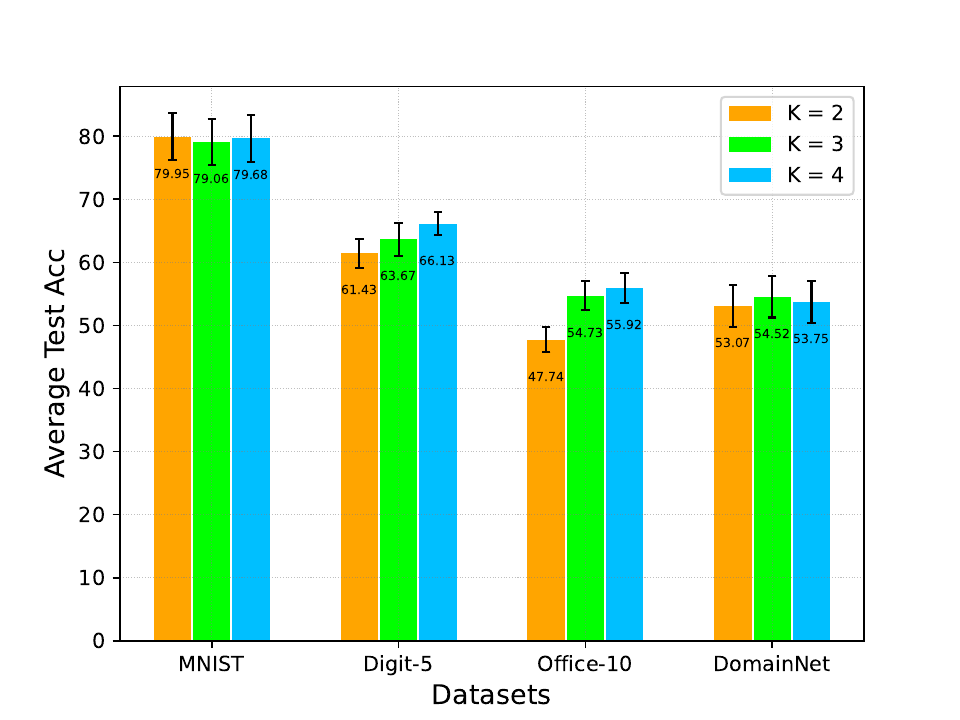}}
\caption{Average test accuracy (\%) under the label distribution skew with Dir(0.05) on four baseline datasets with $K$ ranging from 2 to 4.}
\label{selectK}
\end{figure}

\begin{figure*}[t]
\centering
\subfigure[MNIST]{\includegraphics[scale=0.24]{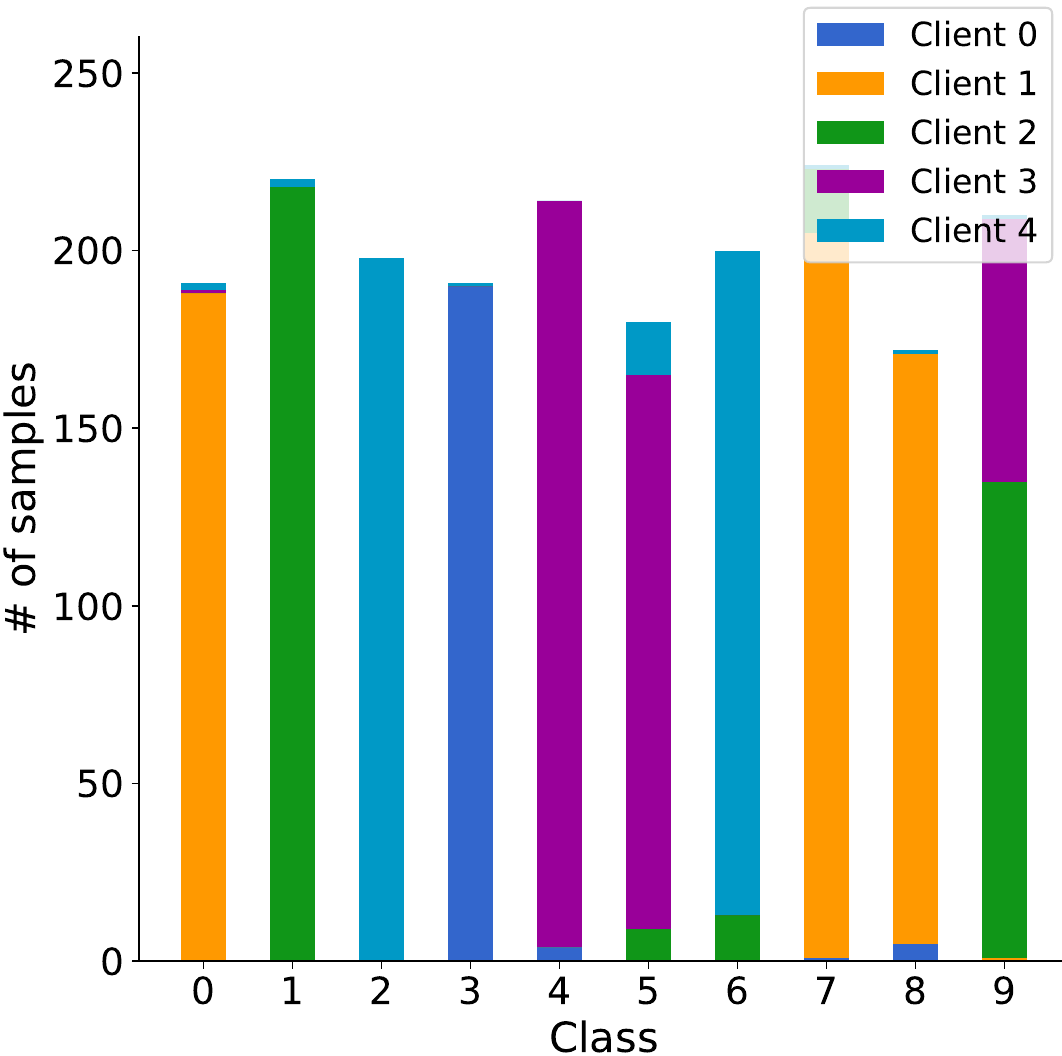}}
\subfigure[Digit-5]{\includegraphics[scale=0.24]{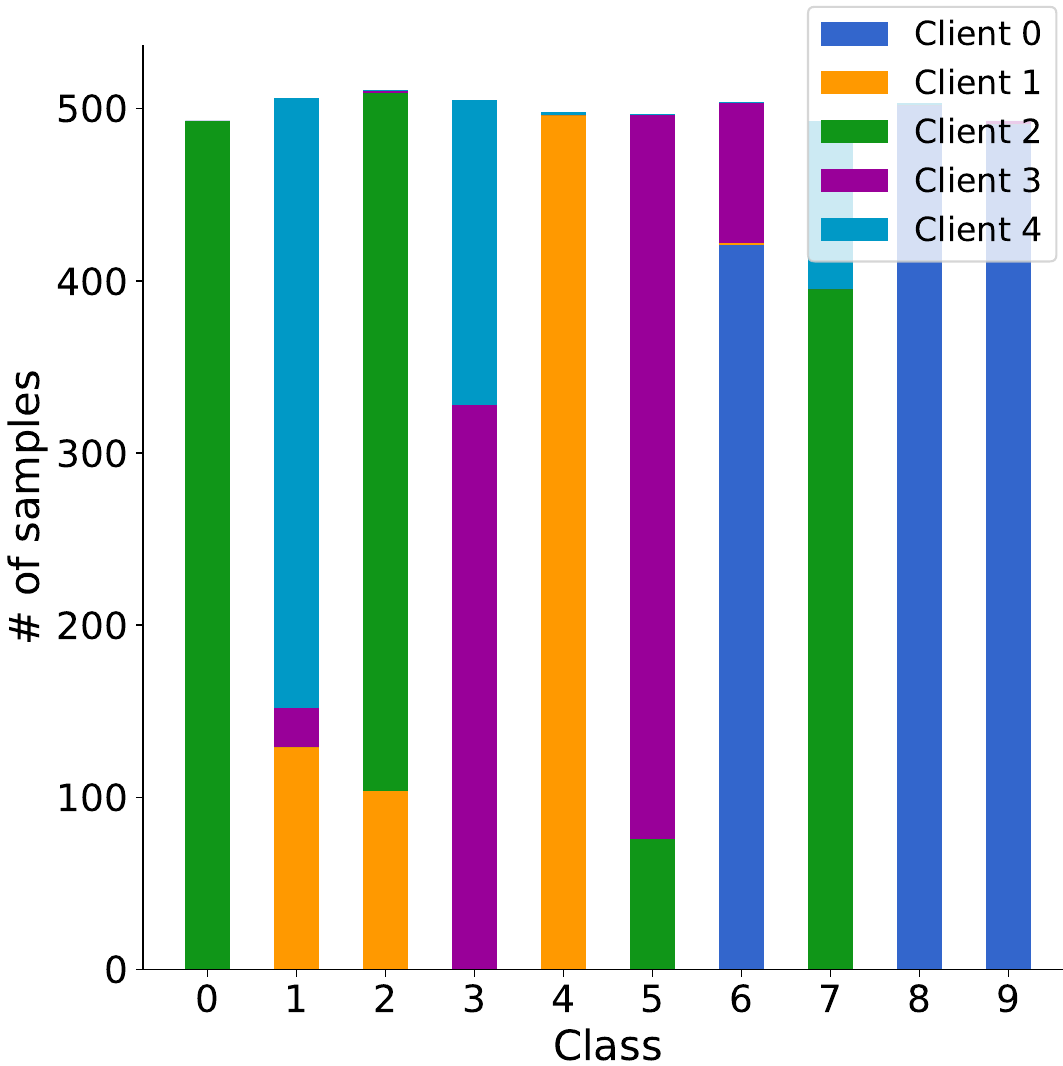}}
\subfigure[Office-10]{\includegraphics[scale=0.24]{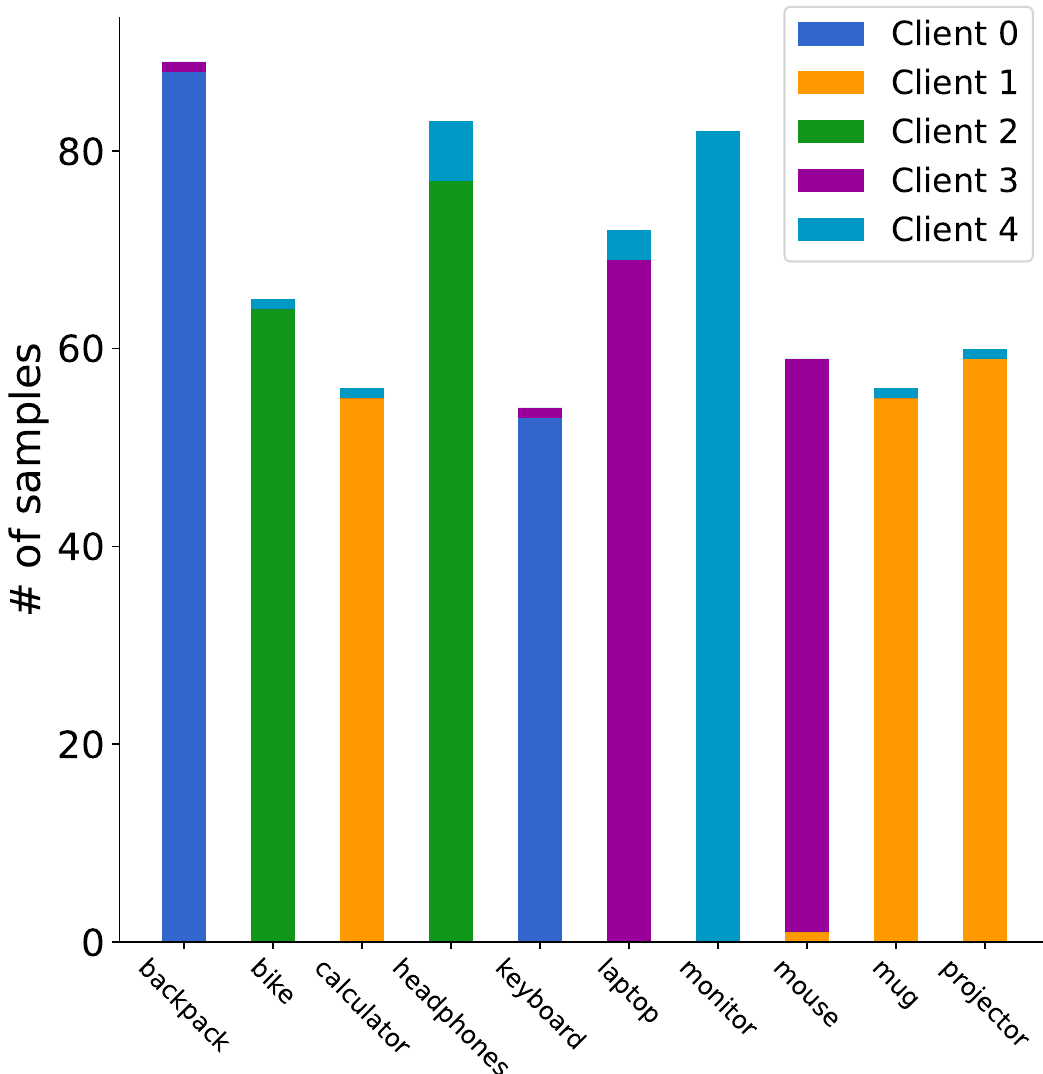}}
\subfigure[DomainNet]{\includegraphics[scale=0.24]{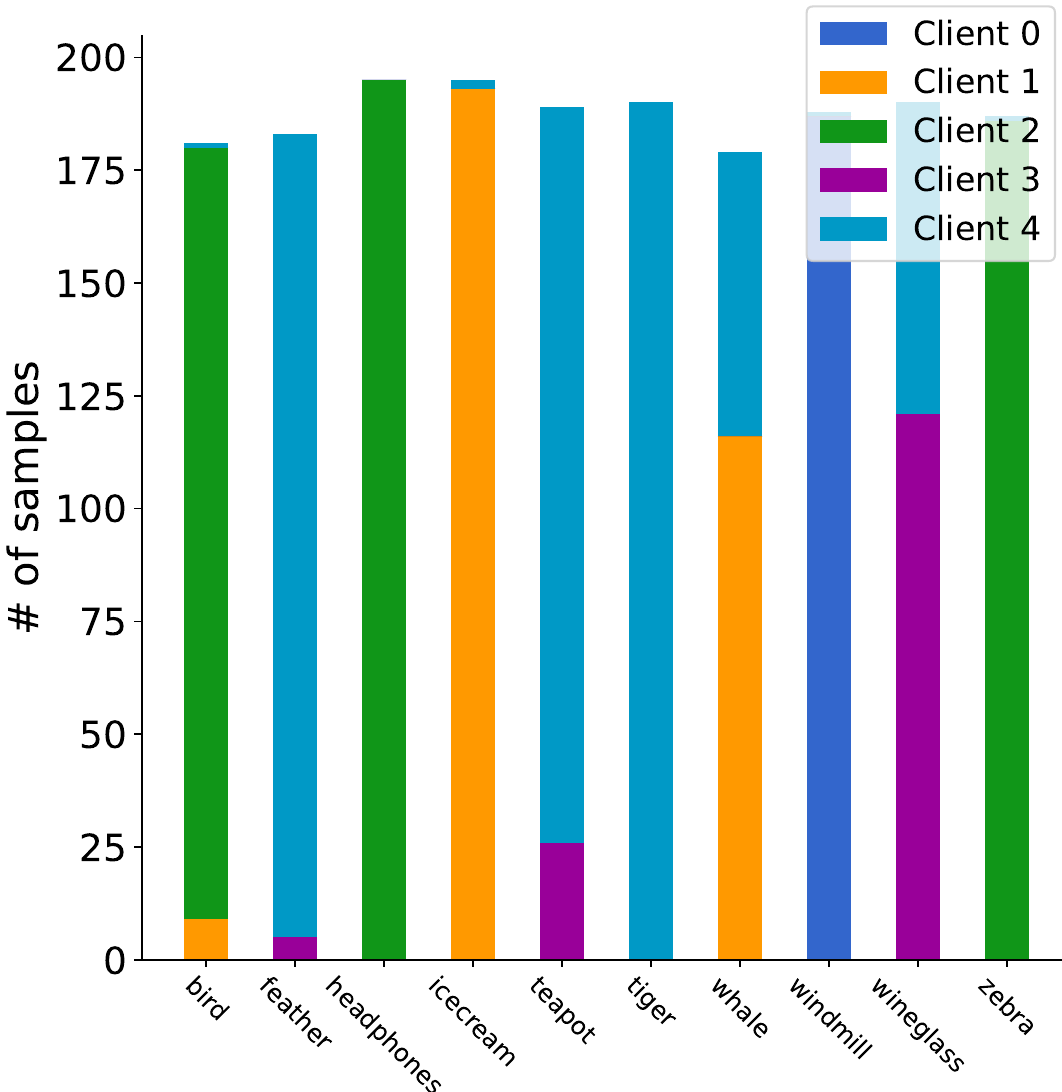}}
\caption{ {Illustration of the \textit{label non-IID} setting on MNIST, Digit-5, Office-10, and DomainNet datasets. The same color represents the same client, and each client samples from their respective dataset according to Dir(0.05). The sampling results have the same feature distribution but different label distribution.}}
\label{label_nonIID}
\end{figure*}

\begin{figure*}[t]
\centering
\subfigure[Digit-5]{\includegraphics[scale=0.25]{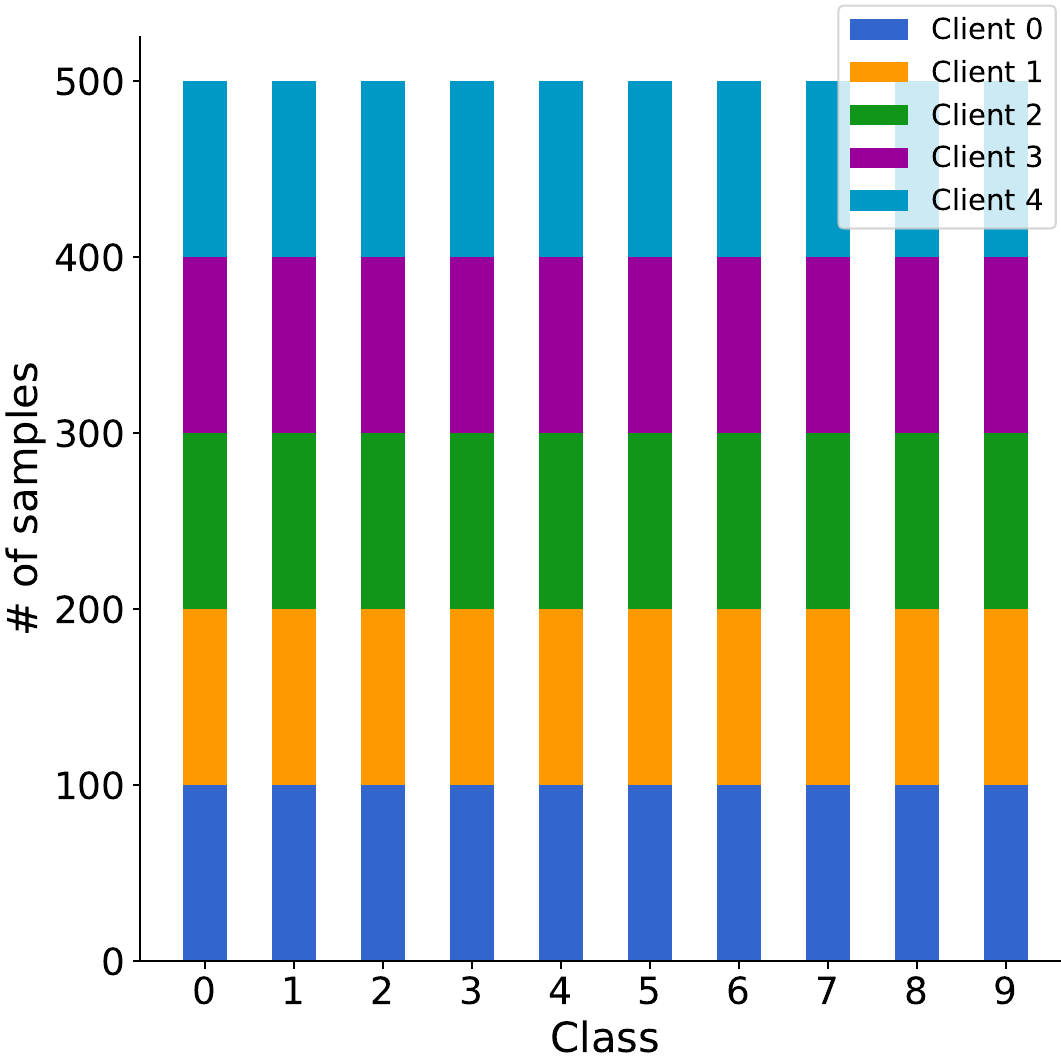}}
\subfigure[Office-10]{\includegraphics[scale=0.25]{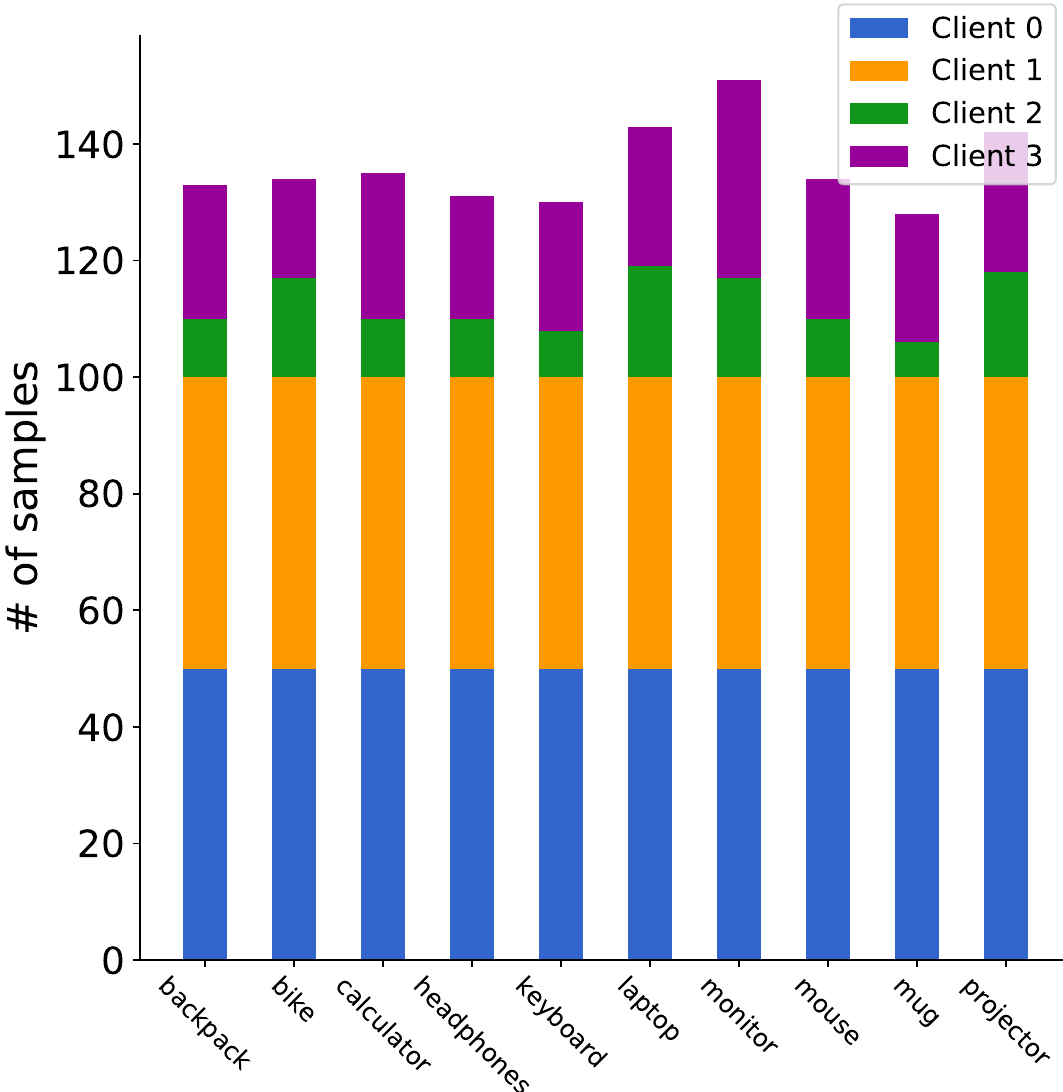}}
\subfigure[DomainNet]{\includegraphics[scale=0.25]{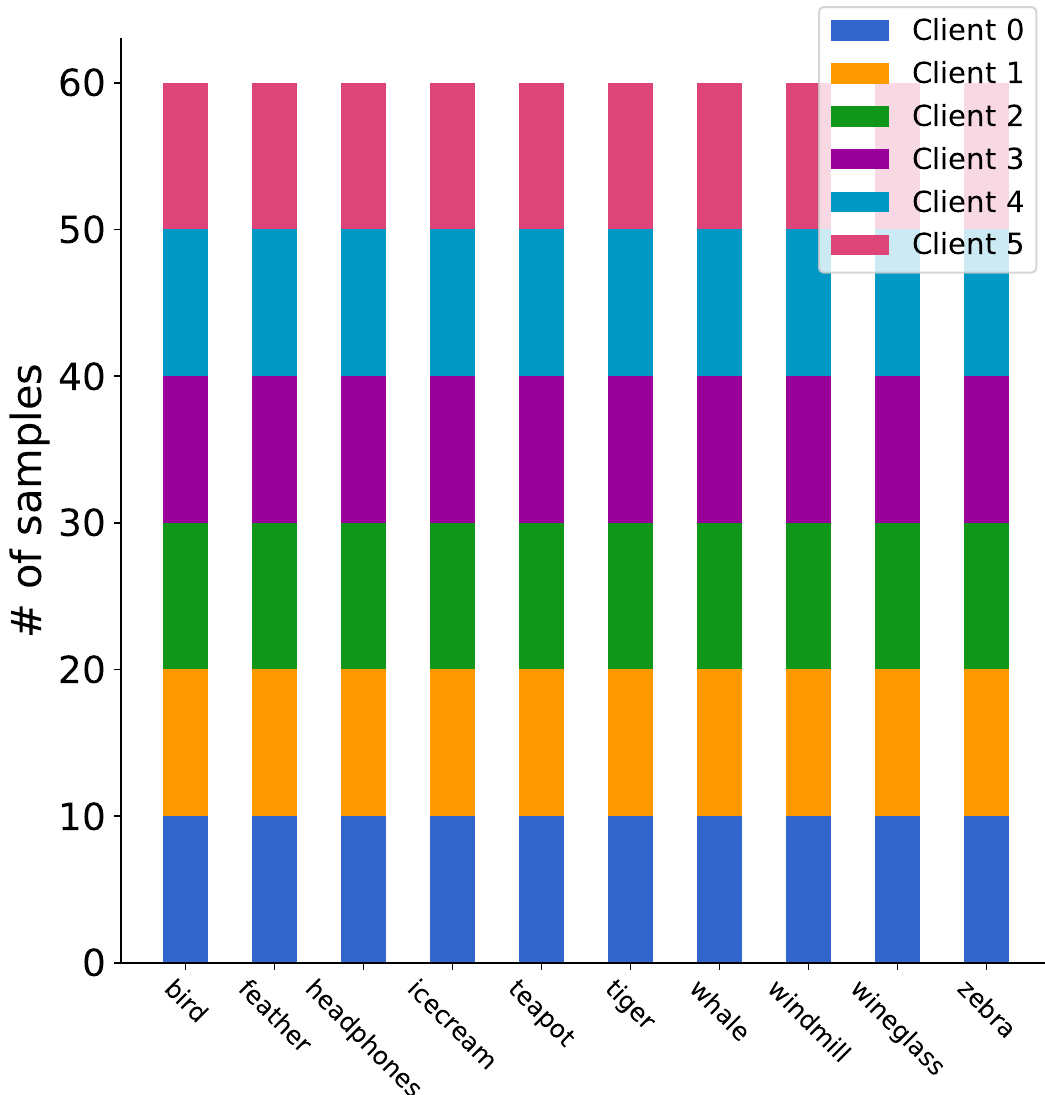}}
\caption{ {Illustration of the \textit{feature non-IID} setting on Digit-5, Office-10, and DomainNet datasets. The different color represents clients with different datasets, and the visualization result shows that each client shares the same label distribution but has different feature distribution.}}
\label{feature_nonIID}
\end{figure*}

\vspace{-0.1cm}
\subsection{Implementation Details}
\label{ID}
We investigate two different non-IID settings to mimic non-IID scenarios : (i) feature distribution skew: clients have the same label distributions but different feature distributions, (ii) label distribution skew: clients have different label distributions but the same feature distribution, which is simulated by Dirichlet distribution Dir($\alpha$) \cite{yurochkin2019bayesian}. Here, the more skewness among clients is, the smaller the value $\alpha$ is, and vice versa. The label distribution skewness $\alpha$ is set to 0.05 for all federated training algorithms unless explicitly specified. 

We compare our proposed method with popular FL algorithms including Local where local models are updated in each global round without any communication with others, FedAvg \cite{mcmahan2017communication}, FedProto \cite{tan2022fedproto}, and FedProx \cite{li2020federated}. We use 5, 4, and 6 clients for Digit-5, Office-10, and DomainNet in the feature distribution skewed setting, respectively. In the label distribution skewed setting, the number of clients for Digit-5, Office-10, and DomainNet is all 5. We use 5 clients for MNIST unless explicitly specified. The size of MNIST for all experiments is 2,000 for simplicity. The visualization results of all datasets with label non-IID and feature non-IID settings are shown in Figure \ref{label_nonIID} and Figure \ref{feature_nonIID}, respectively.

We use PyTorch \cite{paszke2019pytorch} to implement all the baselines. Following~\cite{li2020federated,tan2022federated}, the grid search is used to select the optimal hyperparameters for model training. Specifically, we use the SGD optimizer for all baselines, and the SGD momentum is set to 0.5. The other training parameters are set to be $B$ = 32, $E$ = 1, $\tau$ = 0.07, $\eta$ = 0.01 with decay rate = 0.95, which denotes local batch size, local epochs, temperature, learning rate, and  {learning rate decay per iteration}, respectively.

\vspace{-0.2cm}
\subsection{Choosing $K$}
\label{CK}
The selection of $K$ is performed in the feature space of samples in different datasets. Intuitively, the feature space of samples from different datasets is different; thus the number of $K$ is associated with specific datasets.  {Here, we adopt a similar way as \cite{rippel2015metric} to maintain a uniform value for $K$ across classes for simplicity, although this value may vary for different classes. Note that since we focus on the fixed-prototype strategy in this paper, we initially explore a dynamic prototype-based scheme for completeness. Specifically,} we study the effect of different $K \in \{2,3,4\}$ under label distribution skew through heuristic selection and then apply the most appropriate $K$ value to all experiments under different non-IID settings.  {Similar heuristic selection method applied in the feature space has been utilized in many papers~\cite{gupta2017local,deuschel2021multi,qiao2023framework}.} We run three trials with different random seeds and report the average test accuracy on validation datasets, as shown in Figure \ref{selectK}. The number of communication rounds is set to 60 for all datasets. It can be found that the appropriate $K$ values for MNIST, Digit-5, Office-10, and DomainNet are 2, 4, 4, and 3, respectively. The hyperparameters used are presented in Table \ref{tab:parameters}.

\begin{table}
\centering
\caption{Top-1 average test accuracy (\%) of MP-FedCL and other baselines on Digit-5 under \textit{feature non-IID} setting. \\}
\label{tab:feature non-IID}
\resizebox{\linewidth}{!}{%
\begin{tabular}{l|ccccc}
    \toprule
    \textbf{Method} & Local & FedAvg & FedProx & FedProto &\textbf{MP-FedCL} \\ 
    \midrule
    MNIST  & \multicolumn{1}{l}{47.33(8.96)} & \multicolumn{1}{l}{53.00(4.32)} & \multicolumn{1}{l}{72.00(5.89)} & \multicolumn{1}{l}{88.00(2.83)}  & \multicolumn{1}{l}{\textbf{91.37}(0.39)}\\
    SVHN  & 16.67(2.05)  & 18.67(1.25)  & 22.33(3.30) & 25.00(1.41) & \textbf{26.80}(2.65)  \\
    USPS  & 60.33(1.25)   & 54.67(4.99)    & 71.67(5.73)   & 91.67(1.25)    & \textbf{93.93}(0.65)  \\
    Synth & 26.33(6.13)  & 36.00(2.94)  & 49.33(4.71)  & 56.67(1.25)   & \textbf{61.00}(0.49)   \\
    MNIST-M & 20.33(3.30) & 32.00(4.55) & 43.67(4.11) & 48.67(3.68) & \textbf{51.20}(1.77) \\
    \midrule
    Average & \multicolumn{1}{l}{34.20(4.34)} & \multicolumn{1}{l}{38.87(3.61)} & \multicolumn{1}{l}{51.80(4.75)} & \multicolumn{1}{l}{62.00(2.08)} & \multicolumn{1}{l}{\textbf{64.86}(1.19)} \\
    \bottomrule
    \end{tabular}
}
\end{table}

\begin{table}
\centering
\caption{ {Top-1 average test accuracy (\%) of MP-FedCL and other baselines on various datasets under \textit{label non-IID} setting.}}
\label{tab:label non-IID}
\resizebox{\linewidth}{!}{%
\begin{tabular}{l|llll|cl} 
\toprule
\textbf{Method}   & MNIST & Digit-5 & Office-10 & DomainNet  &  {{\begin{tabular}[c]{@{}c@{}}\# of Comm \\ Rounds \end{tabular}}}\\ 
\midrule
FedAvg   &  66.40(2.89)& 29.66(1.57)   &  24.00(1.55) & 23.61(4.35) &  {110}\\
FedProx  & 64.85(1.88) & 28.51(2.13)   &  21.74(1.13) & 22.78(5.66) &  {100} \\
FedProto & 33.27(1.74) & 60.84(1.79)   & 39.79(3.15) & 36.02(5.23) &  {100} \\
 {
DP-FedCL} &  {78.49(3.09)} &  {53.94(3.38)} &  {37.22(1.75)} &  {47.62(1.78)} &  {60}\\
SP-FedCL & 79.44(3.48)  & 57.91(3.17) & 43.11(4.30) & 35.44(4.25) &  {60} \\
\textbf{MP-FedCL} & \textbf{79.95}(3.76)   & \textbf{67.15}(2.33)  & \textbf{59.07}(3.41)   &  \textbf{52.12}(5.02) &  {60} \\
\bottomrule
\end{tabular}}
\end{table}

\vspace{-0.2cm}
\subsection{Accuracy Comparison}
\label{AC}
 {In this section, given that the number of $K$ may vary across different classes, we attempt to explore another feature clustering method for feature clustering, called density-based spatial clustering of applications with noise (DBSCAN)~\cite{ester1996density}, which is different from \textit{k-means} as it does not require specifying the number of clusters beforehand. It automatically determines the number of clusters based on the data's density and has been applied in the image field~\cite{he2021image,hou2016dsets,shen2016real,mehdi2019improving,harisinghaney2014text}. Here, we combine the DBSCAN clustering method with our proposal that each client uses DBSCAN for clustering instead of \textit{k-means}, terming it DP-FedCL, and compare it with MP-FedCL and other baselines.}

Specifically, we compare our method with the baselines under the \textit{feature non-IID} and \textit{label non-IID} settings where \textit{feature non-IID} means the feature distribution is skewed while the label distribution is IID, and \textit{label non-IID} means the label distribution is skewed while the feature distribution is IID. For the sake of fair comparison, we conduct experiments with the same hyperparameters, run three trials, and report the mean and standard derivation. The best results are shown in \textbf{bold}.
 
Table \ref{tab:feature non-IID} reports the average test accuracies of our method and baselines in the mean(std) format under the \textit{feature non-IID} setting. The results indicate that MP-FedCL achieves higher test accuracy and smaller standard deviation compared to most cases, with an improvement of approximately 4.6\% over the second-highest test accuracy. In addition, Table \ref{tab:label non-IID} presents our results and those of the baselines under the \textit{label non-IID} setting. In this setting, we use SP-FedCL, which refers to our training strategy that employs a single prototype (i.e., $K$ = 1). Specifically, we do not use any clustering algorithm in Eq. \ref{cluster}, but instead take the mean value of the feature space belonging to the same label as the prototype of the label, which is also known as a single prototype. The results show that our method enjoys relatively significant advantages over almost all other baselines, and about at least 10.4\% improvement in test accuracy compared to most cases. Moreover, since the cost of communication has been considered to be the bottleneck of FL, we also report the number of communication rounds for each algorithm in Table \ref{tab:label non-IID} accordingly.  {It can be seen that our proposal and other methods based on our strategy only need relatively few communication rounds compared to others. For DP-FedCL, the only difference from MP-FedCL is that it dynamically selects the number of clusters using the DBSCAN algorithm on each local client. The results indicate that clustering using \textit{k-means} and assigning an equal number of prototypes to each class outperforms the adopted dynamic selection approach. We conjecture that while dynamically selecting the number of clusters might lead to varying prototype assignments for each class, it can also introduce an imbalance in class representations, resulting in biased model learning.}

\begin{figure}[t]
\centering
\subfigure[Digit-5]{\includegraphics[scale=0.6]{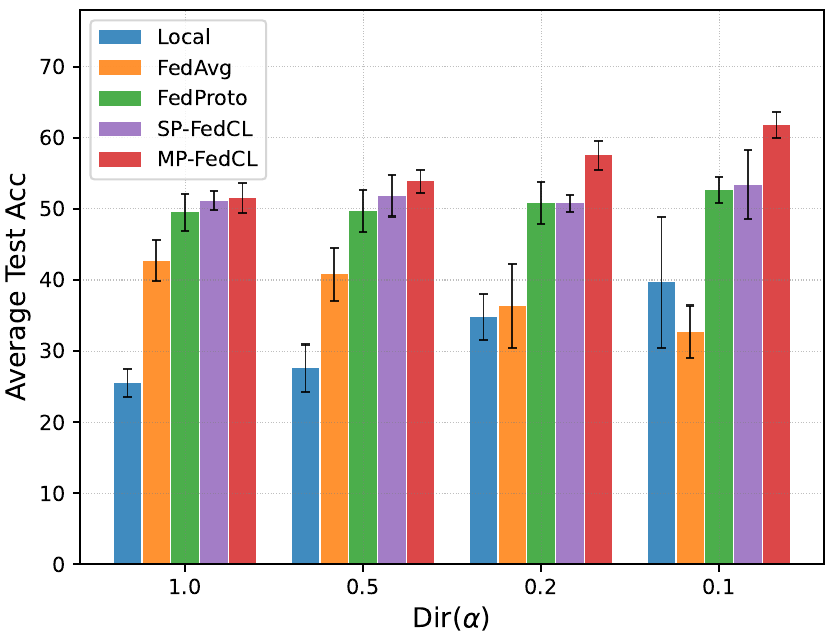}}
\subfigure[Office-10]{\includegraphics[scale=0.6]{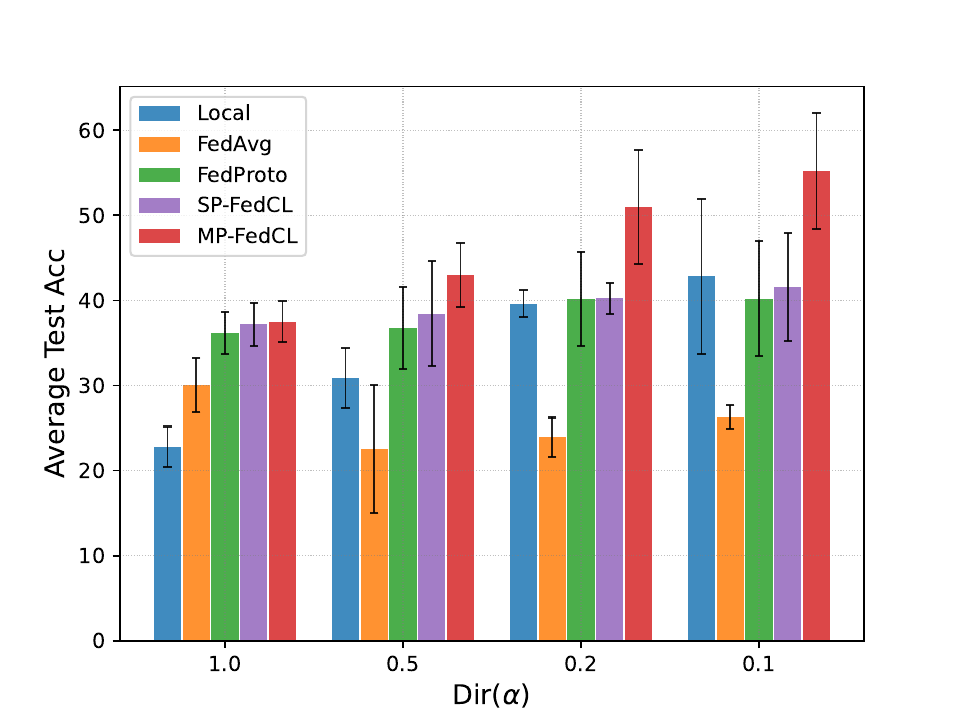}}
\caption{Illustration of the average test accuracy (\%) on Digit-5 and Office-10 under different levels of \textit{label non-IID} settings.}
\label{diff_alpha_acc}
\end{figure}
\vspace{-0.1cm}
\subsection{Robustness Comparison}
\label{RC}
\textbf{Different degrees of label non-IID.} Considering that labeling non-IID is a common challenge in FL, designing an algorithm that is robust to various heterogeneous data is crucial for deploying the FL algorithm in real applications. Therefore, we compare our method with several baselines under different levels of label heterogeneity to verify the robustness of different algorithms to data heterogeneity. As shown in Figure \ref{diff_alpha_acc}, the $\alpha$ decreases from 1.0 to 0.1, controlling the degree of label heterogeneity, which means that labels are distributed more and more heterogeneously among clients. We report the average test accuracy for Digit-5 under different heterogeneities in Figure \ref{diff_alpha_acc} (a). Note that since our proposal requires only a small number of iterations, therefore the number of communication rounds for MP-FedCL and SP-FedCL is set to 60, and  100 for others. The results show that our proposal outperforms all approaches under different heterogeneous settings in terms of test accuracy, and enjoys a relatively small deviation compared to those in most cases. 

 {Specifically, in comparison to the performance under all heterogeneity settings, our method achieves at least an 8.8\% and 2.0\% increase in accuracy compared to the popular baseline FedAvg and state-of-the-art FedProto, respectively. To highlight the advantage of multi-prototype learning over single-prototype learning, we conduct experiments using the same learning strategy as the former, but with only a single prototype, which we refer to as SP-FedCL. The results show that our method consistently outperforms SP-FedCL by approximately 2.0\% in average test accuracy across various scenarios. These results underscore the effectiveness of our multi-prototype learning approach in FL, leading to improved robustness compared to both single-prototype and state-of-the-art methods. Interestingly, in some scenarios (such as $\alpha = 0.1$), the performance of FedAvg is lower than that of Local, which shows that FedAvg does not always perform well in dealing with heterogeneous scenarios. It would be an intriguing research direction to explore the synergies between local optimization and federated optimization, allowing us to harness the advantages offered by both approaches.}

Additionally, in Figure \ref{diff_alpha_acc} (b), we present similar results for Office-10 under various degrees of heterogeneity. The findings further demonstrate that, in the majority of cases, our proposed method outperforms all other approaches in terms of test accuracy, highlighting its advantage in dealing with data heterogeneity. Specifically, compared to SP-FedCL, FedProto, and FedAvg, our method shows improvements of at least 0.3\%, 1.4\%, and 7.5\%, respectively, across different label non-IID settings. However, it is worth noting that almost all methods, including ours, encounter significant deviations under extremely heterogeneous settings like Dir(0.1) and Dir(0.2). This can be attributed to the relatively limited number of training samples available in Office-10 compared to Digit-5, which may lead to higher fluctuations in performance. Moreover, it should be noted that similar results are observed in this setting, with FedAvg performing worse than Local in most cases.

\begin{figure}[t]
\centering
\subfigure[ {MNIST}]{\includegraphics[scale=0.6]{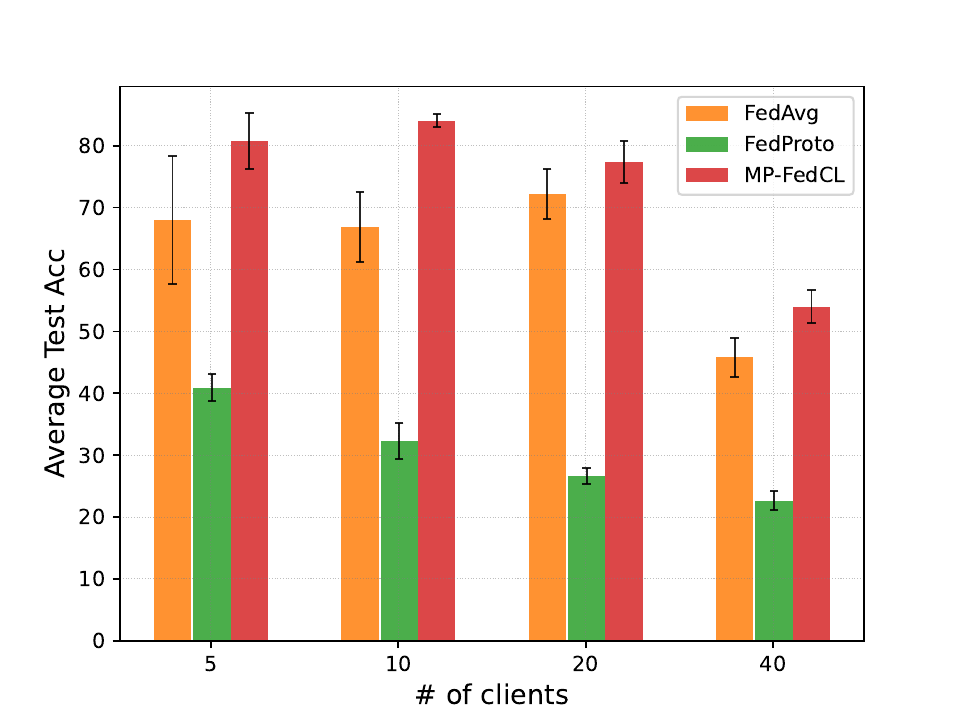}}
\subfigure[Digit-5]{\includegraphics[scale=0.6]{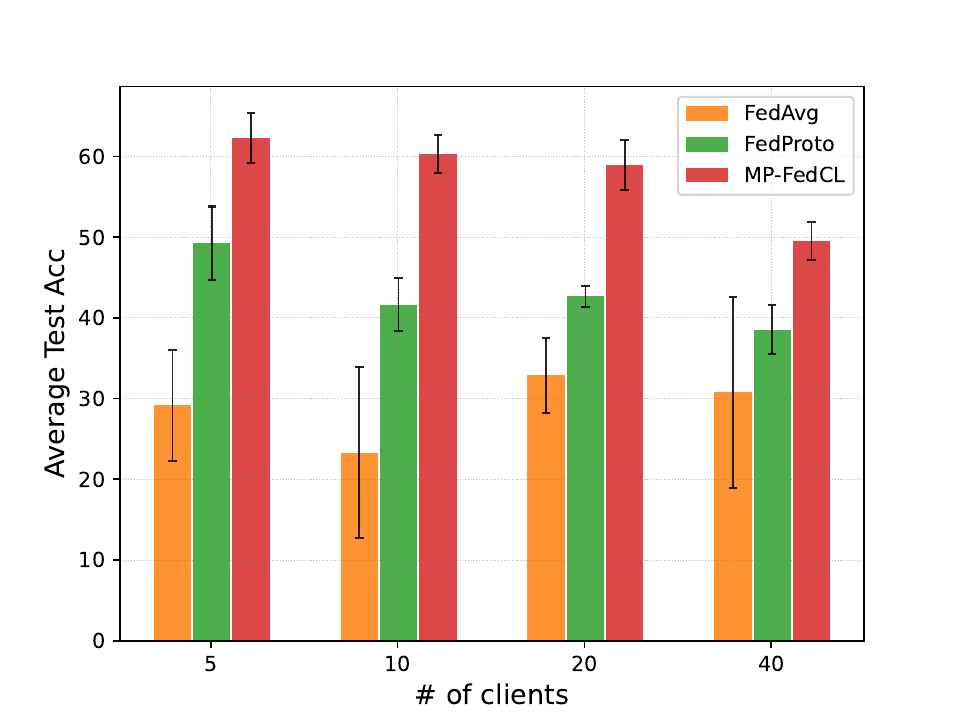}}
\caption{ {Illustration of the average test accuracy (\%) on MNIST and Digit-5 based on Dir(0.1) with the varying numbers of clients.}}
\label{diff_num_acc}
\end{figure}

\begin{figure}[t]
\centering
\subfigure[Digit-5]{\includegraphics[scale=0.6]{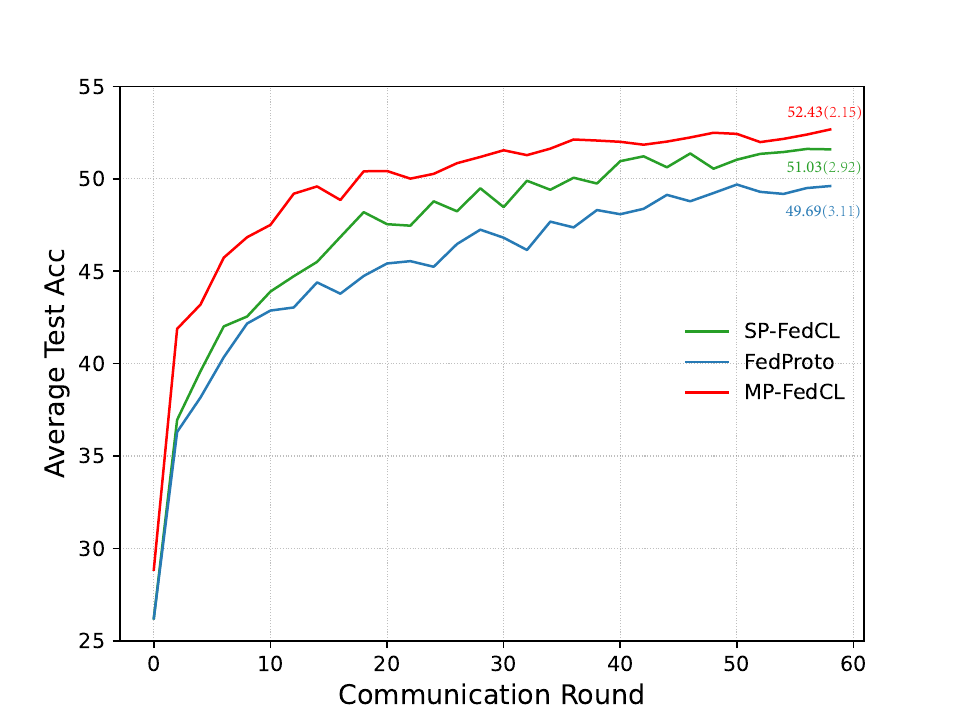}}
\subfigure[Office-10]{\includegraphics[scale=0.6]{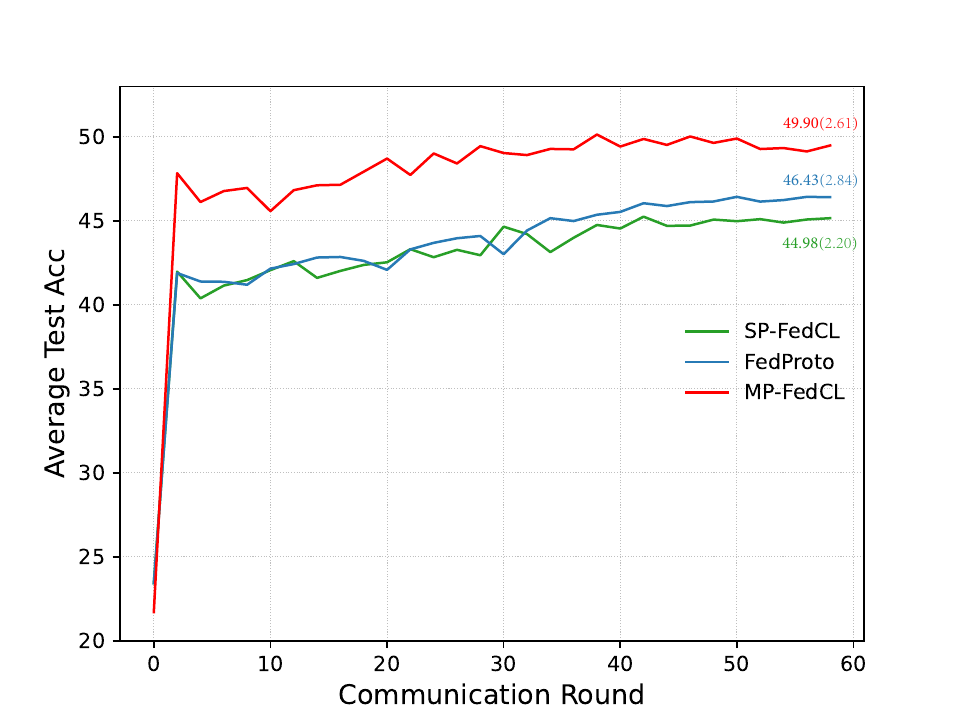}}
\caption{ {Illustration of the average test accuracy (\%) in each global round on Digit-5 and Office-10, respectively. The upper subfigure result is reported under \textit{label non-IID} with Dir(1.0), and the below is reported under \textit{feature non-IID}.}}
\label{comm_effici_acc}
\end{figure}

\textbf{Different numbers of clients.} 
In addition to exploring the robustness of our proposed algorithm to different levels of data heterogeneity, we also aim to investigate its robustness to varying numbers of participating clients in the FL setting.  {As depicted in Figure \ref{diff_num_acc}, we demonstrate the robustness of our proposed method by evaluating its performance across an increasing number of clients, ranging from 5 to 40, with labels distribution following Dir(0.1).} The average test accuracy for MNIST and Digit-5 under the various number of clients is shown in Figure \ref{diff_num_acc} (a) and Figure \ref{diff_num_acc} (b), respectively.  {The illustration demonstrates that our proposed method exhibits a distinct advantage over FedAvg in terms of test accuracy across different numbers of clients. Moreover, our approach maintains a relatively small deviation compared to FedAvg in most cases, further highlighting its robustness against various numbers of clients.}

 {Specifically, regarding the results of MNIST displayed in Figure \ref{diff_num_acc} (a), our proposed method achieves a remarkable increase of at least 7.2\% in test accuracy compared to the baseline FedAvg when varying the number of clients. It is interesting to note that, compared to both FedAvg and MP-FedCL, FedProto exhibits inferior performance across all numbers of participating clients. Moreover, the performance of FedProto decreases as the number of clients increases, and its peak performance is only about half of that of MP-FedCL. This observation strongly suggests that the approach of not transmitting the model parameters, as adopted in FedProto, may not be suitable for models trained from scratch. Similarly, in the case of Digit-5, as shown in Figure \ref{diff_num_acc} (b), our proposal consistently outperforms both FedAvg and FedProto in terms of test accuracy across different numbers of clients. To be precise, our method demonstrates an improvement of at least 18.73\% and 10.99\% compared to FedAvg and FedProto, respectively. Furthermore, it is worthwhile to mention that our proposed method exhibits lower accuracy variance compared to the other methods in the majority of cases, whereas FedAvg experiences higher fluctuations. This observation could be attributed to the inherent complexity of the feature space in Digit-5, which is likely more intricate compared to the MNIST dataset. Consequently, the variations between clients in Digit-5 become more pronounced, leading to the observed differences in performance for FedAvg. In contrast, prototype-based schemes such as FedProto and MP-FedCL show more robust results (in terms of accuracy variation) across different numbers of participating clients in the FL setting. This indicates that an approach based on prototypes successfully handles heterogeneity to a certain extent.}

\begin{table}
\centering
\caption{ {Top-1 average test accuracy (\%) of MP-FedCL and other baselines on Digit-5 and Office-10 under \textit{label non-IID} with Dir(0.5) and \textit{feature non-IID}.}}
\label{tab:commun_effici}
 {
\resizebox{\linewidth}{!}{%
\begin{tabular}{l|lll|cl} 
\toprule
 {\textbf{Dataset}} &  {\textbf{Method}} &  {\textit{Label non-IID}} &  {\textit{Feature non-IID}} &  {{\begin{tabular}[c]{@{}c@{}}\# of Comm \\ Rounds \end{tabular}}}\\ 
\midrule
   &  Local & 27.60(3.33)   & 33.60(3.25)  & 0 \\
   &  FedAvg & 40.80(3.71)   & 43.13(3.67)  & 130 \\
Digit-5 
 & FedProto & 49.71(2.98)   & 63.99(0.38) & 130       \\
 & DP-FedCL & 50.37(2.51)  & 63.49(2.02) & 60       \\
 & SP-FedCL & 51.86(2.92)   & \textbf{65.49(1.14)} & 60       \\
 & \textbf{MP-FedCL} & \textbf{53.87(1.62)}   & 65.08(0.91)  & 60       \\
\midrule
   &  Local & 30.89(3.50)   & 21.47(3.72)  & 0 \\
   &  FedAvg & 22.51(7.50)   &  30.17(2.15)  & 130 \\
Office-10
  & FedProto & 36.75(3.50)   & 46.58(2.82)  & 130       \\
  & DP-FedCL & 35.74(5.33)   & 42.77(1.03) & 60       \\
 & SP-FedCL & 38.44(6.15)   & 48.88(3.59) & 60       \\
 & \textbf{MP-FedCL} & \textbf{42.98(3.79)}  & \textbf{49.70(1.49)} & 60       \\
\bottomrule
\end{tabular}}}
\end{table}

\subsection{Communication Efficiency Comparison}
\label{CEC}
 {As FL involves training models across distributed clients, ensuring a fast and efficient convergence rate is of great importance. A faster convergence rate implies that the participating clients can converge to the optimal with higher accuracy in fewer communication rounds, reducing the overall communication and computational overhead. Therefore, in this section, we evaluate the convergence rate of our proposal and conduct a comprehensive investigation with other benchmarks.}

 {
Table \ref{tab:commun_effici} presents the top-1 average test accuracy (\%) of different methods, including Local, FedAvg, FedProto, DP-FedCL, SP-FedCL, and MP-FedCL, on two datasets: Digit-5 and Office-10 and the corresponding communication rounds. The evaluation is performed under the condition of \textit{label non-IID} with Dir(0.5) and \textit{feature non-IID}. From the results in Table~\ref{tab:commun_effici}, we observe that MP-FedCL consistently achieves the highest top-1 average test accuracy and fewer communication rounds compared to other methods in most cases. Specifically, Local achieves the lowest accuracy since it does not participate in FL rounds (0 communication rounds). The methods based on federated training such as FedAvg and FedProto, have shown improved performance, but they require a high number of communication rounds (130 communication rounds). In contrast,  MP-FedCL consistently outperforms the other federated methods in most cases and only requires about half the number of communication rounds (60 communication rounds) of FedAvg, showcasing its capability to learn from distributed data efficiently. In other words, achieving similar performance to the baselines MP-FedCL requires fewer computational resources. Moreover, we observe a similar phenomenon as in Table~\ref{tab:label non-IID} that DP-FedCL is still lower than the single-prototype scheme and our proposal in this setting. However, it is worthwhile to note that other schemes for dynamically selecting prototypes may potentially outperform our current approach, which leaves for our future work.}

\begin{table}
\centering
\caption{ {Top-1 average test accuracy (\%) of MP-FedCL and other baselines on Digit-5 and Office-10 under \textit{label non-IID} with Dir(0.5) and \textit{feature non-IID}.}}
\label{tab:fair_commun_effici}
 {
\resizebox{\linewidth}{!}{%
\begin{tabular}{l|llc|lcl} 
\toprule
\textbf{Dataset} & \textbf{Method} & \textit{Label non-IID} & \textbf{{\begin{tabular}[c]{@{}c@{}}Training \\ Time (s)\end{tabular}}} & \textit{Feature non-IID} & \textbf{{\begin{tabular}[c]{@{}c@{}}Training \\ Time (s) \end{tabular}}}\\ 
\midrule
   &  FedAvg & 38.17(1.26)  & $\approx$ 600 & 43.00(3.49)  & $\approx$ 600  \\
 & FedProto & 52.67(1.03)  & $\approx$ 600 & 62.00(2.69) & $\approx$ 600    \\
 Digit-5
 & DP-FedCL & 52.42(1.83) & $\approx$ 600 & 64.07(2.07) & $\approx$ 600   \\
 & SP-FedCL & 52.46(1.58) & $\approx$ 600 & \textbf{65.07(1.89)} & $\approx$ 600   \\
 & \textbf{MP-FedCL} & \textbf{52.94(1.68)} & $\approx$ 600 & 65.00(1.47)  & $\approx$ 600 \\
\midrule
   &  FedAvg & 23.59(5.69)  & $\approx$ 380 &  24.04(2.61)  & $\approx$ 380 \\
  & FedProto & 39.11(6.99)  & $\approx$ 380 & 45.34(2.72) & $\approx$ 380        \\
Office-10
  & DP-FedCL & 35.98(5.95) & $\approx$ 380 & 41.83(2.14) & $\approx$ 380      \\
 & SP-FedCL & 39.60(5.89) & $\approx$ 380 & 45.20(3.56) & $\approx$ 380      \\
 & \textbf{MP-FedCL} & \textbf{43.59(5.54)} & $\approx$ 380& \textbf{47.38(1.48)} & $\approx$ 380  \\
\bottomrule
\end{tabular}}}
\end{table}

 {Table~\ref{tab:fair_commun_effici} shows the top-1 average test accuracy of MP-FedCL and other baselines on Digit-5 and Office-10 under \textit{label non-IID} with Dir(0.5) and \textit{feature non-IID}. All algorithms are compared under similar training times. The results show that MP-FedCL achieves comparable or superior performance compared to baselines. In other words, MP-FedCL can still achieve competitive performance under approximately the same computational resources.} In addition, we compare the average test accuracy of our proposed MP-FedCL with FedProto and SP-FedCL per global round during training based on different kinds of heterogeneous settings (including \textit{feature non-IID} and \textit{label non-IID}) after several runs with different random seeds, as shown in Figure \ref{comm_effici_acc}. Both subfigures demonstrate that our method outperforms them in terms of test accuracy and communication efficiency. Specifically, we evaluate our proposal with FedProto and SP-FedCL on Digit-5 under \textit{label non-IID} with Dir(1.0). It shows that our method outperforms theirs by approximately 2.74\% and 1.40\% in test accuracy, respectively, and also leads them the way in convergence rate. Similar results are available in Figure \ref{comm_effici_acc} (b), which is the test accuracy of Office-10 with \textit{feature non-IID}. It demonstrates that our proposal still outperforms theirs by 3.47\% and 4.92\% in terms of test accuracy, and they lag behind us by a significant margin in terms of convergence rate. This indicates the potential advantages of using multi-prototype learning in handling heterogeneous tasks in FL, and we believe that multi-prototype learning based on architectures such as transformers will be a promising research direction in the future.  {Moreover, it can be expected that combining strategies aimed at accelerating \textit{k-means} calculation and other privacy-preserving techniques with our strategy will further make contributions to the FL community.}

\section{Conclusion}
\label{sec:conclusion}
In this paper, we have introduced multi-prototype learning into the federated learning framework, and at the same time, a contrastive learning strategy is applied to multi-prototype learning to make full use of multi-prototype knowledge. First, a clustering-based multi-prototype calculation approach has been proposed. In order to better leverage the knowledge from each client, we then have proposed a contrastive learning strategy that encourages clients to learn class-related knowledge from others in each global round through multi-prototype exchange, while reducing the absorption of class-unrelated knowledge. Further, we have introduced a multi-prototype-based model inference strategy into FL. This strategy also provides the potential for fast model training in distributed edge networks, as only a small amount of training is required when adding a new client, to compare its prototype with the trained prototypes in the global prototype pool for fast and correct predictions. Finally, extensive experiments on four datasets demonstrate that our proposal has a robust performance against label and feature heterogeneity in terms of test accuracy and communication efficiency. Compared to several baselines, our test accuracy improves by about 4.6\% and 10.4\% under feature and label non-IID, respectively. 

\bibliographystyle{IEEEtran}
\bibliography{bib_local}

\appendices
\section*{appendix}
\subsection{Convergence Analysis}
\label{appex:CA}
We provide a convergence analysis for MP-FedCL. We add a subscript to the local objective defined in Eq. \ref{local_loss} indicating the number of global iterations and give the following assumptions following the literature \cite{li2020federated, tan2022fedproto, wu2022node, chai2021fedat}.
\vspace{-0.1cm}
\begin{ass} 
\label{ass1}
\textbf{Convex, $\zeta$-Lipschitz, and $L_1$-smooth}.\\
The local loss objective $\mathcal{L}_i(\omega)$ is convex, $\zeta$-Lipschitz, and $L_1$-smooth for each client $i$, 
\begin{equation}
\begin{aligned}
 \Vert\mathcal{L}_i(\omega_{t_1}) - \mathcal{L}_i(\omega_{t_2}) \Vert  \leq \zeta \Vert  \omega_{t_1} -  \omega_{t_2} \Vert, \\
 \quad \Vert \nabla\mathcal{L}_i(\omega_{t_1}) -\nabla \mathcal{L}_i(\omega_{t_2}) \Vert  \leq L_1 \Vert \omega_{t_1} -  \omega_{t_2} \Vert,\\
\forall t_1, t_2 >0, i \in \{1,2,\dots, N\}.
\end{aligned}
\end{equation}
\end{ass}

Based on Assumption \ref{ass1}, the definition of $\mathcal{L}_i(\omega)$, and triangle inequality, it is easy to prove that $\mathcal{L}_i(\omega)$ is convex, $\zeta$-Lipschitz, and $L_1$-smooth.

\begin{ass} 
\label{ass2}
\textbf{$\delta$-local dissimilarity}.\\
Each local loss function $\mathcal{L}_i(\omega_t)$ is $\delta$-local dissimilar at $\omega_t$, i.e.,\\
\begin{equation}
\small
\begin{aligned}
{\Bbb E}_{i \sim \mathcal{D}_i}[ \Vert \nabla \mathcal{L}_i(\omega_{t}) \Vert^2 ] \leq \Vert \nabla \mathcal{L}(\omega_{t}) \Vert^2 \delta^2, \\
\forall t>0, i \in \{1,2,\dots, N\}.
\end{aligned}
\end{equation}
\end{ass}
${\Bbb E}_{i \sim \mathcal{D}_i}[\bigcdot]$ denotes the expectation over client $i$. $\nabla \mathcal{L}(\omega_{t})$ is the global gradient at the $t$-th global round, which can be defined as $\nabla \mathcal{L}(\omega_{t}) = \frac{1}{D_i} \sum_{i\in \mathcal{D}_i} \nabla \mathcal{L}_i(\omega_{t})$. This assumption is to ensure the bounded similarity between the local model and the global model, thus ensuring the convergence of the model.

\begin{ass} 
\label{ass4}
\textbf{Bounded Gradient}.\\
The expectation of the stochastic gradient of each client $i$ is bounded by $G$, i.e.,
\begin{equation}
\small
{\Bbb E}[{\Vert \nabla \mathcal{L}_i(\omega_{t};\xi_{t}) \Vert}] \leq G, \forall i \in \{1,2,\dots, N\}.
\end{equation}
\end{ass}

\begin{ass}
\label{ass5}
\textbf{Local embedding $L_2$-Lipschitz continuous}. \\
Each local embedding function is $L_2$-Lipschitz continuous, that is,
\begin{equation}
\small
\begin{aligned}
\| f_{e,i}(\omega_{e,t_1})&- f_{e,i}(\omega_{e,t_2})\| \leq L_2\| \omega_{e,t_1} -\omega_{e,t_2} \|, \\
&\forall t_1, t_2 >0, i \in \{1,2,\dots, N\}.
\end{aligned}
\end{equation}
\end{ass}

Assumptions from \ref{ass1} to \ref{ass4} are standard, and have been made in different variants by previous works \cite{li2020federated, wu2022node, tan2022fedproto, chai2021fedat}. Further, the similar assumption defined in Assumption \ref{ass5} can be found in their works \cite{tan2022fedproto}. The main purpose of this assumption is to control the changing rate of the local embedding function, making it easier to study its behaviour and properties. 

\begin{lemma}
\label{lemma1}
Let assumptions \ref{ass1} and \ref{ass2} hold. For an arbitrary client, the expected decrement in the local loss between two consecutive global rounds satisfies
\begin{equation}
\small
\begin{aligned}
 \mathcal{L}(\omega_{t+1}) & \leq   \mathcal{L}(\omega_{t}) - (\eta  - \frac{L_1\eta^2}{2})\delta^2 \Vert \nabla \mathcal{L}(\omega_{t})) \Vert^2,
\end{aligned}
\end{equation}
where $\eta$ is the learning rate of SGD.
\end{lemma}

The proof of {Lemma} \ref{lemma1} is presented in Appendix\ref{apex1}. Lemma \ref{lemma1} provides a bound on how rapid the decrease of the local loss for each client can be expected before the multi-prototype aggregation in each global round. 

\begin{lemma}
\label{lemma2}
Let Assumptions \ref{ass4} and \ref{ass5} hold. After the multi-prototype calculation is completed, the loss function of an arbitrary client can be bounded as:
\begin{equation}
\small
\begin{aligned}
\mathbb{E}\left [\mathcal{L}(\omega_{t+1+\delta_t})\right]
 &\leq \mathcal{L}(\omega_{t+1}) + \frac{1}{N} \sum_{i \in \mathcal{D}_i} \sum_{i=1}^{N}\frac{A_{p}L_2\eta G}{\tau} v_i^{t+1}, 
\end{aligned}
\end{equation}
where $p \in P(y)$, $A_p$ is the size of labels distinct from $p$.
\end{lemma}

The proof of {Lemma} \ref{lemma2} is presented in Appendix\ref{apex2}. Lemma \ref{lemma2} provides a bound on how rapid the decrease of the local loss for each client can be expected after the multi-prototype aggregation in each global round. 

\begin{theorem}
\label{thm1}
Let Assumptions \ref{ass1} to \ref{ass5} hold. Assume a fixed unit local updates between two consecutive global communication rounds. For an arbitrary client, after every communication round, we have,
{
\begin{equation} \label{eq: th1}
\small
\begin{aligned}
\mathbb{E}\left [\mathcal{L}(\omega_{t+1+\delta_t})\right]
&\leq \mathcal{L}(\omega_{t}) - (\eta  - \frac{L_1\eta^2}{2})\delta^2 \Vert \nabla \mathcal{L}(\omega_{t})) \Vert^2 \\
&+ \frac{1}{N} \sum_{i \in \mathcal{D}_i} \sum_{i=1}^{N}\frac{A_{p}L_2\eta G}{\tau} v_i^{t+1}.
\end{aligned}
\end{equation}
}
\end{theorem}
The proof of {Theorem} \ref{thm1} is presented in Appendix\ref{apex3}. Theorem \ref{thm1} expresses the deviation bound of an arbitrary client's local objective after each global round. The expected decrease in loss per global round can be achieved through the selection of an appropriate $\eta$, thus guaranteeing convergence.

\begin{corollary} \label{colla1}
\noindent The loss function $\mathcal{L}$ of an arbitrary client monotonously decreases in every communication round when
\begin{equation}
\small
\eta \leq \frac{2}{L_1} - \frac{2\sum_{i \in \mathcal{D}_i}\sum_{i=1}^{N}A_{p}L_{2}G v_i^{t+1}}{L_1 N \tau \delta^2 \Vert \nabla \mathcal{L}(\omega_{t}) \Vert^2}.
\end{equation}
Therefore, the loss function reaches convergence.
\end{corollary}

The proof of {Corollary} \ref{colla1} is presented in Appendix\ref{apex3}. Corollary \ref{colla1} is intended to ensure that the one-round expected deviation of $\mathcal{L}$ is negative, thus achieving convergence of the loss function, which can guide the selection of an appropriate learning rate $\eta$ to ensure convergence.

\begin{theorem}
\label{thm2}
Let Assumptions \ref{ass1} to \ref{ass5} hold. For an arbitrary client, MP-FedCL achieves a upper bound after $T$ global rounds, when
\begin{equation} \label{eq: th2}
\small
\begin{aligned}
\eta \leq \frac{2}{L_1} - \frac{2\sum_{i \in \mathcal{D}_i}\sum_{i=1}^{N}A_{p}L_{2}G v_i}{L_1 N \tau \delta^2 \epsilon}.
\end{aligned}
\end{equation}
\end{theorem} 
The proof of {Theorem} \ref{thm2} is presented in Appendix\ref{apex4}. Theorem \ref{thm2} provides the convergence rate, indicating that achieving a tighter bound $\epsilon$ requires more communication rounds $T$ and smaller learning rate $\eta$, which can guide the selection of learning rate and communication round.

\subsection{Proof of Lemma \ref{lemma1}}
\label{apex1}
Since this lemma is for arbitrary clients, the client notation $i$ is omitted. Based on the $L_1$-smooth in Assumption \ref{ass1} of $\mathcal{L}(\omega_t)$ and applying Taylor expansion \cite{wu2022node}, we have
\begin{align*}
\label{a1}
\mathcal{L}(\omega_{t+1}) \leq  \mathcal{L}(\omega_{t}) +  \langle \nabla \mathcal{L}(\omega_{t}), \omega_{t+1} - \omega_{t} \rangle + \frac{L_1}{2} \Vert \omega_{t+1} - \omega_{t} \Vert ^2,
\tag{A1}
\end{align*}
which implies the following quadratic bound,
\begin{align*}
\label{a2}
\mathcal{L}(\omega_{t+1}) - \mathcal{L}(\omega_{t}) \leq   \langle \nabla \mathcal{L}(\omega_{t}), \omega_{t+1} - \omega_{t} \rangle + \frac{L_1}{2} \Vert \omega_{t+1} - \omega_{t} \Vert ^2,
\tag{A2}
\end{align*}

Based on the local updates calculated in Eq. \ref{sgd} and taking expectations, we have a bounded $\Vert \omega_{t+1} - \omega_{t} \Vert^2$ as
\begin{align*}
\label{a3}
 \Vert \omega_{t+1} -\omega_{t}\Vert^2 
 & =  (\mathbb{E}_{i \backsim \mathcal{D}_i} \left[ \Vert \omega_{t+1} - \omega_{t}\Vert \right])^2 \\
 & = \eta^2  (\mathbb{E}_{i \backsim \mathcal{D}_i} \left[ \Vert  \nabla  \mathcal{L}_i(\omega_{t})) \Vert  \right])^2\\
  &\stackrel{(a)}\leq  \eta^2  \mathbb{E}_{i \backsim \mathcal{D}_i} \left[ \Vert  \nabla \mathcal{L}_i(\omega_{t})) \Vert^2 \right] \\
  &\stackrel{(b)} \leq  \eta^2  \Vert \nabla \mathcal{L}(\omega_{t})) \Vert^2 \delta^2,
 \tag{A3}
\end{align*}
where (a) holds because of Cauchy-Schwarz inequality $(\mathbb{E}  \Vert \mathbf{X} \Vert)^2 \leq  \mathbb{E}  \Vert \mathbf{X}^2 \Vert $, and (b) follows from the bounded dissimilarity Assumption \ref{ass2}.

Further, $\langle \nabla \mathcal{L}(\omega_{t}), \omega_{t+1} - \omega_{t} \rangle$ can be reformulated as:
\begin{align*}
\label{a4}
\langle \nabla \mathcal{L}(\omega_{t}), \omega_{t+1} - \omega_{t} \rangle 
& = -\eta \mathbb{E}_{i \backsim \mathcal{D}_i} [\langle \nabla \mathcal{L}(\omega_{t}), \omega_{t+1} - \omega_{t} \rangle] \\
& \stackrel{(a)} =  -\eta \mathbb{E}_{i \backsim \mathcal{D}_i}\left[\langle \nabla \mathcal{L}(\omega_{t}),  \nabla \mathcal{L}_i(\omega_{t})  \rangle \right] \\
&\stackrel{(b)} \approx -\eta  \mathbb{E}_{i \backsim \mathcal{D}_i} \left[ \Vert  \nabla \mathcal{L}_i(\omega_{t})) \Vert^2 \right] \\
& \stackrel{(c)}\leq -\eta \Vert \nabla \mathcal{L}(\omega_{t})) \Vert^2 \delta^2,
 \tag{A4}
\end{align*}
where (a) holds because of the definition for SGD optimization calculated in Eq. \ref{sgd}, (b) follows from $\nabla \mathcal{L}(\omega_{t})$  is calculated by aggregating over local updates across distributed clients, (c) follows from the bounded dissimilarity Assumption \ref{ass2}, and ${\Bbb E}_{i \sim \mathcal{D}_i}[\bigcdot]$ denotes the expectation over client $i$.

\noindent Substituting (\ref{a3}) and (\ref{a4}) into (\ref{a2}), then
\begin{align*}
\label{a5}
\begin{aligned}
\mathcal{L}(\omega_{t+1}) - \mathcal{L}(\omega_{t}) \leq  \frac{L_1\eta^2}{2}  \Vert \nabla \mathcal{L}(\omega_{t}) \Vert^2 \delta^2 -\eta \Vert \nabla \mathcal{L}(\omega_{t}) \Vert^2 \delta^2.
\end{aligned}
\tag{A5}
\end{align*}
Then (A5) can be reformulated into
\begin{align*}
\label{a6}
\begin{aligned}
\small
\mathcal{L}(\omega_{t+1}) \leq   \mathcal{L}(\omega_{t}) - (\eta  - \frac{L_1\eta^2}{2})\delta^2 \Vert \nabla \mathcal{L}(\omega_{t})) \Vert^2.
 \end{aligned}
\tag{A6}
\end{align*}
\subsection{Proof of Lemma \ref{lemma2}}
\label{apex2}
Let Assumptions \ref{ass4} and \ref{ass5} hold, we suppose that $t+1+\delta_t$ denotes the training process between $t+1$ and $t+2$ global round, that is, the multi-prototype calculation process before the next global iteration starts, meaning that $\delta_t \in (0, 1)$. Then, the local loss function can be reformulated as \cite{tan2022fedproto}:
\begin{align*}
\small
\label{a7}
\mathcal{L}(\omega_{t+1+\delta_t}) &= \mathcal{L}(\omega_{t+1}) + \mathcal{L}(\omega_{t+1+\delta_t}) - \mathcal{L}(\omega_{t+1}) \\
& = \mathcal{L}(\omega_{t+1}) + \mathcal L_R(\omega_{t+1+\delta_t}) - \mathcal L_R(\omega_{t+1}).
\tag{A7}
\end{align*}

\vspace{0.1cm}
\noindent Let $Z = \exp(v_i \bigcdot u_{p} / \tau) / \sum_{a \in [C]}\exp(v_i \bigcdot u_{a} /\tau)$ for short, then
\begin{small}
\begin{align*}
&\mathcal L_R(\omega_{t+1+\delta_t}) - \mathcal L_R(\omega_{t+1})  \\
&= \frac{1}{KN|P(y)|} \sum_{i \in \mathcal{D}_i} \sum_{i=1}^{N} \sum_{k=1}^{K} \sum_{p\in P(y)}\log\left({Z_{t+1}} / {Z_{t+1+\delta_t}}\right).
\label{a8}
\tag{A8}
\end{align*}
\end{small}

\noindent Here, we first calculate $Z_{t+1} / Z_{t+1+\delta_t}$ in (A8) for simplicity, then
\begin{small}
\begin{align*}
    &{Z_{t+1}} / {Z_{t+1+\delta_t}} \\
    & = \frac{\exp(v_i^{t+1} \bigcdot u_{p}^{t+1} / \tau)}{\sum_{a \in [C]}\exp(v_i^{t+1} \bigcdot u_{a}^{t+1} /\tau)} / \frac{\exp(v_i^{t+1} \bigcdot u_{p}^{t+2} / \tau)}{\sum_{a \in [C]}\exp(v_i^{t+1} \bigcdot u_{a}^{t+2} /\tau)} \\
    &= \frac{\exp(v_i^{t+1} \bigcdot u_{p}^{t+1} /\tau)}{\exp(v_i^{t+1} \bigcdot u_{p}^{t+2} /\tau)} \cdot \frac{\sum_{a \in [C]}\exp(v_i^{t+1} \bigcdot u_{a}^{t+2} /\tau)}{\sum_{a \in [C]}\exp(v_i^{t+1} \bigcdot u_{a}^{t+1} /\tau)} \\
    &= \exp\left(\frac{v_i^{t+1} \bigcdot (u_{p}^{t+1} - u_{p}^{t+2})}{\tau}\right) \cdot \sum_{a \in [C]}\exp(\frac{v_i^{t+1} \bigcdot \left( u_{a}^{t+2} - u_{a}^{t+1}\right) }{\tau}).
\label{a9}
\tag{A9}
\end{align*}
\end{small}

\noindent Taking $\log$ operation of both sides of the above equation, we have
\allowdisplaybreaks[1]
\begin{footnotesize}
\begin{align*}
    &\log\left({Z_{t+1}} / {Z_{t+1+\delta_t}}\right) \\
    &= \frac{v_i^{t+1} \bigcdot (u_{p}^{t+1} - u_{p}^{t+2})}{\tau} + \log \left(\sum_{a \in [C]}\exp(\frac{v_i^{t+1} \bigcdot \left( u_{a}^{t+2} - u_{a}^{t+1}\right) }{\tau})\right) \\
    &\stackrel{(a)} \leq \frac{v_i^{t+1} \bigcdot (u_{p}^{t+1} - u_{p}^{t+2})}{\tau} + \sum_{a \in [C]}\frac{v_i^{t+1} \bigcdot \left( u_{a}^{t+2} - u_{a}^{t+1}\right) }{\tau} \\
    &= \frac{v_i^{t+1}}{\tau} \bigcdot \left[(u_{p}^{t+1} - u_{p}^{t+2}) + \sum_{a \in [C]}\left( u_{a}^{t+2} - u_{a}^{t+1}\right)\right] \\
    &\stackrel{(b)}= \frac{v_i^{t+1}}{\tau} \bigcdot \left[(u_{p}^{t+1} - u_{p}^{t+2})  + (u_{p}^{t+2} - u_{p}^{t+1})+\sum_{a \in [C] / p}\left( u_{a}^{t+2} - u_{a}^{t+1}\right)\right] \\
    &= \frac{v_i^{t+1}}{\tau} \bigcdot \sum_{a \in [C] / p}\left( u_{a}^{t+2} - u_{a}^{t+1}\right) \\
    &\stackrel{(c)} \leq \frac{v_i^{t+1}}{\tau} \bigcdot \sum_{a \in [C] / p} \Vert u_{a}^{t+2} - u_{a}^{t+1} \Vert \\
    &\stackrel{(d)} \approx \frac{v_i^{t+1}}{\tau} \bigcdot \sum_{a \in [C] / p} \Vert v_{a}^{t+2} - v_{a}^{t+1} \Vert \\
    &\stackrel{(e)} = \frac{v_i^{t+1}}{\tau} \bigcdot \sum_{a \in [C] / p} \Vert f_{e,i}(\omega_{e,t+2};\boldsymbol x_{i,a}) - f_{e,i}(\omega_{e,t+1};\boldsymbol x_{i,a}) \Vert \\
    &\stackrel{(f)} \leq A_p \frac{v_i^{t+1}}{\tau} \bigcdot L_2\| \omega_{e,t+2} -\omega_{e,t+1} \| \\
    &\stackrel{(g)} \leq \frac{A_{p}L_2}{\tau} v_i^{t+1} \bigcdot \| \omega_{t+2} -\omega_{t+1} \| \\
    &\stackrel{(h)} \leq \frac{A_{p}L_2\eta}{\tau} v_i^{t+1} \bigcdot \Vert  \nabla  \mathcal{L}_i(\omega_{t})) \Vert.
\label{a10}
\tag{A10}
\end{align*}
\end{footnotesize}

\noindent Substituting (\ref{a10}) and (\ref{a8}) into (\ref{a7}), and taking expectations of random variable $\xi$ on both sides, then
\begin{footnotesize}
\begin{align*}
\mathbb{E}\left [\mathcal{L}(\omega_{t+1+\delta_t})\right] 
&\leq \mathcal{L}(\omega_{t+1}) +\frac{1}{N} \sum_{i \in \mathcal{D}_i} \sum_{i=1}^{N} \frac{A_{p}L_2\eta}{\tau} v_i^{t+1} \bigcdot \mathbb{E}\left [\Vert  \nabla  \mathcal{L}_i(\omega_{t})) \Vert \right]\\
 &\stackrel{(i)} \leq \mathcal{L}(\omega_{t+1}) + \frac{1}{N} \sum_{i \in \mathcal{D}_i} \sum_{i=1}^{N}\frac{A_{p}L_2\eta G}{\tau} v_i^{t+1}, 
\label{a11}
\tag{A11}
\end{align*}
\end{footnotesize}

\noindent where (a) follows from Jensen's Inequality, (b) holds because $a \in [C]/p$ is created to represent the set of labels distinct from $p$, (c) holds by applying triangle inequality, (d) holds from the considering that the equation still approximately holds after clustering where we consider $\frac{1}{K}\sum_{k=1}^{K}(\cdot) = 1$ for simplicity, (e) follows from Eq. \ref{feature_extrac}, (f) holds because of Assumption \ref{ass5} and the size of $a \in [C]/p$ is represented as $A_p$, (g) holds because the fact that $\omega_{e,t}$ is a subset of $\omega_{t}$, (h) follows from the definition of SGD in Eq. \ref{sgd}, (i) holds because of Assumption. \ref{ass4}.

\subsection{Proof of Theorem \ref{thm1} and Corollary \ref{colla1}}
\label{apex3}
\vspace{0.2cm}
Taking the expectation of $\omega$ on both sides of Lemma \ref{lemma1} and Lemma \ref{lemma2}, and summing them, we have
\begin{footnotesize}
\begin{align*}
\mathbb{E}\left [\mathcal{L}(\omega_{t+1+\delta_t})\right]
&\leq \mathcal{L}(\omega_{t}) - (\eta  - \frac{L_1\eta^2}{2})\delta^2 \Vert \nabla \mathcal{L}(\omega_{t})) \Vert^2 \\
&+ \frac{1}{N} \sum_{i \in \mathcal{D}_i} \sum_{i=1}^{N}\frac{A_{p}L_2\eta G}{\tau} v_i^{t+1}.
\label{a12}
\tag{A12}
\end{align*}
\end{footnotesize}

\noindent Then, in order to ensure $- (\eta  - \frac{L_1\eta^2}{2})\delta^2 \Vert \nabla \mathcal{L}(\omega_{t})) \Vert^2 + \frac{1}{N} \sum_{i \in \mathcal{D}_i} \sum_{i=1}^{N}\frac{A_{p}L_2\eta G}{\tau} v_i^{t+1} \leq 0$, we have
\begin{small}
\begin{align*}
\eta \leq \frac{2}{L_1} - \frac{2\sum_{i \in \mathcal{D}_i}\sum_{i=1}^{N}A_{p}L_{2}G v_i^{t+1}}{L_1 N \tau \delta^2 \Vert \nabla \mathcal{L}(\omega_{t}) \Vert^2}.
\label{a13}
\tag{A13}
\end{align*}
\end{small}

Theorem \ref{thm1} and Corollary \ref{colla1} are proved, and the convergence of $\mathcal{L}$ holds.

\subsection{Proof of Theorem \ref{thm2}}
\label{apex4}
Theorem \ref{thm2} is proven by combing the deviation bound in Theorem \ref{thm1} with the result in [\cite{tan2022fedproto}, Theorem 2], taking the expectation of the result in Theorem \ref{thm1} with respect to $\omega$ on both sides, we have,
\begin{align*}
\sum_{t=0}^{T-1}({\Bbb E}[\mathcal{L}(\omega_{t})] - {\Bbb E} [\mathcal{L}(\omega_{t+1+\delta_t})]) \leq \Delta,
\tag{A14}
\end{align*}
\noindent where $\mathcal{L}(\mathbf{\omega}_{0})-\mathcal{L}( \mathbf{\omega}^*) = \Delta$, and $\mathcal{L}(\mathbf{\omega}_{0})$ denotes the loss at the first global round, and $\mathcal{L}( \mathbf{\omega}^*)$ represents the optimal local model parameters that minimize $\mathcal{L}( \mathbf{\omega})$.

\noindent Then, considering a total of $T$ global communication rounds, we have,
\begin{align*}
\begin{aligned}
& \frac{1}{T}\sum_{t=0}^{T-1}\mathbb{E}\left [\Vert\nabla \mathcal{L}(\omega_{t}) \Vert^2\right] 
\leq \\& \frac{\frac{1}{TN}\sum_{i \in \mathcal{D}_i} \sum_{i=1}^{N}\sum_{t=0}^{T-1}\frac{A_{p}L_2\eta G}{\tau} v_i^{t+1} + \frac{\Delta}{T}}{(\eta - \frac{L_1 \eta^2}{2})\delta^2}.
\end{aligned}
\label{a15}
\tag{A15}
\end{align*}

\noindent Given some $\epsilon > 0$ which indicates that there exists a positive constant $\epsilon$ that is greater than the right side of the inequality given in (\ref{a15}). Then
\begin{align*}
 \epsilon &\geq \frac{\frac{1}{TN}\sum_{i \in \mathcal{D}_i} \sum_{i=1}^{N}\sum_{t=0}^{T-1}\frac{A_{p}L_2\eta G}{\tau} v_i^{t+1} + \frac{\Delta}{T}}{(\eta - \frac{L_1 \eta^2}{2})\delta^2}.
\label{a16}
\tag{A16}
\end{align*}

\noindent Let $\frac{1}{T}\sum_{t=0}^{T-1}v_i^{t+1} = v_i $, that is,
\begin{align*}
T &\geq \frac{N\Delta}{(\eta - \frac{L_1 \eta^2}{2})N\delta^2\epsilon - \sum_{i \in \mathcal{D}_i}\sum_{i=1}^{N}\frac{A_{p}L_2\eta G}{\tau} v_i }.
\label{a17}
\tag{A17}
\end{align*}

\noindent Therefore, after $T$ communication rounds, we have
\begin{align*}
& \frac{1}{T}\sum_{t=0}^{T-1}\mathbb{E}\left [\Vert\nabla \mathcal{L}(\omega_{t}) \Vert^2\right] \leq \epsilon,
\label{a18}
\tag{A18}
\end{align*}
where a smaller $\epsilon$ means a tighter upper bound, which requires more communication rounds $T$ in (\ref{a17}), when
\begin{align*}
\eta \leq \frac{2}{L_1} - \frac{2\sum_{i \in \mathcal{D}_i}\sum_{i=1}^{N}A_{p}L_{2}G v_i}{L_1 N \tau \delta^2 \epsilon}.
\label{a19}
\tag{A19}
\end{align*}

Finally, Theorem \ref{thm2} convergence rate is proved.

\end{document}